\PassOptionsToPackage{table}{xcolor}
\documentclass[11pt]{article}

\usepackage[final]{acl}

\usepackage{times}
\usepackage{latexsym}
\usepackage{amsmath}
\usepackage{booktabs}

\usepackage[T1]{fontenc}

\usepackage[utf8]{inputenc}

\usepackage{microtype}

\usepackage{inconsolata}

\usepackage{graphicx}
\usepackage{fancyvrb}
\usepackage[most]{tcolorbox} 

\usepackage{hyperref}
\usepackage{url}
\usepackage{xspace} 
\usepackage{multirow}
\usepackage{array}
\usepackage[table]{xcolor}
\usepackage{threeparttable, siunitx}
\usepackage{adjustbox} 
\usepackage{makecell}
\usepackage{titletoc}
\usepackage{fvextra} 
\tcbuselibrary{skins,breakable} 
\usepackage{wrapfig}
\usepackage{algorithm}
\usepackage{algpseudocode}
\usepackage{subcaption}
\usepackage{tabularx}
\usepackage{amssymb}
\usepackage{longtable}
\usepackage{rotating}

\DefineVerbatimEnvironment{mdverbatim}{Verbatim}{
  breaklines=true,    
  breakanywhere=true, 
  obeytabs=true,      
  tabsize=2,         
  showspaces=false,
  showtabs=false,
  fontsize=\scriptsize
}

\newtcolorbox{mdbox}[2][]{
  enhanced,
  breakable,        
  colback=gray!5,     
  colframe=black,  
  boxrule=0.8pt,   
  arc=2mm,   
  colbacktitle=gray!12, 
  coltitle=black, 
  fonttitle=\bfseries, 
  title={#2}, 
  verbatim,  
  #1  
}

\DefineVerbatimEnvironment{caseverbatim}{Verbatim}{
  breaklines=true,    
  breakanywhere=true, 
  obeytabs=true,      
  tabsize=2,         
  showspaces=false,
  showtabs=false,
  fontsize=\scriptsize
}

\newtcolorbox{casebox}[2][]{
  enhanced,
  breakable,        
  colback=blue!5,        
  colframe=blue!50!black,
  boxrule=0.8pt,   
  arc=2mm,   
  colbacktitle=blue!12, 
  coltitle=black, 
  fonttitle=\itshape,
  title={#2}, 
  #1  
}

\title{To Answer or to Abstain: Mitigating Search-Agent Hallucinations via Abstention-Aware Reinforcement Learning}

\author{
\textbf{Fengji Zhang\textsuperscript{1},
Tianyu Fan\textsuperscript{2},
Yuxiang Zheng\textsuperscript{2},
Xinyao Niu\textsuperscript{2},} \\
\textbf{Chengen Huang\textsuperscript{2},
Jacky Keung\textsuperscript{1},
Bei Chen\textsuperscript{2}} \\
\textsuperscript{1}City University of Hong Kong \quad
\textsuperscript{2}Alibaba Group
}

\begin{document}
\maketitle
\begin{abstract}
Recent advances in equipping Large Language Models (LLMs) with search tools and outcome-reward reinforcement learning (RL) have achieved new state-of-the-art results on open-domain QA tasks. However, we argue that current training paradigms harbor a critical vulnerability: they predominantly reward correct answers but fail to penalize fabricated ones when retrieval fails, thereby implicitly exacerbating hallucinations. To address this, we propose Abstention-Aware Reinforcement Learning (\texttt{AWA-RL}), which dynamically shapes the abstention reward utilizing the model's query-specific prior capabilities and continuous on-policy training observations. We also introduce a novel metric, \texttt{RA-F1}, to measure the capability-reliability trade-off. Compared to non-abstaining baselines, \texttt{AWA-RL} boosts absolute precision by up to 10.3\% and overall \texttt{RA-F1} by 2.9\%, with only marginal sacrifice in raw accuracy. These results confirm that \texttt{AWA-RL} successfully yields highly capable and reliable search agents. The code, data, and model weights are publicly available at \url{https://github.com/zfj1998/AWA-RL}.
\end{abstract}

\section{Introduction}
Hallucination remains a long-standing and notoriously difficult challenge in LLMs. It is challenging to ensure the factual grounding of generated output.~\citep{bang2025hallulens, kalai2025language} A prevailing approach to mitigating hallucinations is providing sufficient context to ground the answers. Progressing from early Retrieval-Augmented Generation methods,~\citep{agrawal2024can, bechard2024reducing} recent advancements have introduced more advanced search agents.~\citep{li2025search, team2025tongyi} By leveraging outcome-reward RL, these agents learn to interact with search environments iteratively to arrive at ground-truth answers. 

However, this training paradigm harbors a critical vulnerability: current RL objectives predominantly incentivize models to output the correct answer, but lack a mechanism to reward abstention.~\citep{jin2025search} Consequently, when an agent encounters queries beyond its capability boundaries or fails to retrieve sufficient evidence, it tends to fabricate an answer rather than safely abstain. This ``over-answering'' phenomenon could exacerbate hallucinations.~\citep{song2025hallucination} The critical research question thus arises: \textit{How can we equip search agents with the self-awareness to recognize their capability boundaries and confidently abstain from ungrounded answers?}

Learning to abstain is a non-trivial objective. Existing abstention-related studies \citep{xu2024rejection, song2025hallucination, zhou2026silence, zheng2025enhancing, xu2024reducing} primarily suffer from three major limitations\footnote{Detailed discussions on related works in Appendix~\ref{appendix:related_work}.}: 
(1) \textit{Mismatched Scenarios}: They predominantly focus on single-turn QA rather than multi-turn agentic settings, where capability boundaries dynamically shift based on tool-use and search results.
(2) \textit{Degradation of Capabilities}: Most approaches employ post-hoc pipeline training, which first trains for capability, then tunes for abstention. This disconnected process often severely compromises the model's original search and reasoning abilities.
(3) \textit{Reliance on Heuristics}: They heavily rely on synthesizing new unanswerable datasets, which inevitably introduces human heuristics that fail to generalize across models with varying intrinsic capabilities.

To address these challenges, we propose \textbf{\texttt{AWA-RL}} (\textbf{A}bstention-a\textbf{WA}re \textbf{R}einforcement \textbf{L}earning). The core intuition behind \texttt{AWA-RL} is that the incentive to abstain should not be a universal constant but rather dynamically calibrated to the model's intrinsic problem-solving capability for each specific query. At a high level, we first estimate the model's prior success rate for a given question to establish a query-aware refusal baseline. Second, to counteract the agent's natural bias toward premature ``lazy refusal'', we introduce a non-linear optimistic mapping controlled by a single, explainable ``courage'' factor. Finally, during the RL training phase, we continuously monitor the deviation between the agent's actual abstention behavior and the initial capability estimate, dynamically updating the abstention reward in real time.

Through extensive experiments, we reveal exactly why conventional, intuitive approaches fail to establish a reliable capability-reliability boundary. First, simply manipulating data proportions, such as injecting unanswerable queries or refusal trajectories into the training data, results in highly unpredictable behavior. Second, applying a static refusal reward acts as a blunt instrument that ignores query difficulty variance. Especially in RL from scratch settings, even a minimal static reward triggers catastrophic reward hacking.

In contrast, \texttt{AWA-RL} cleanly overcomes these limitations. Compared to standard non-abstaining RL baselines, our method dramatically boosts absolute precision by up to 10.3\% and improves overall \texttt{RA-F1} by 5.2\%, at the cost of only a marginal drop in raw accuracy. Crucially, empirical results confirm that the ``courage'' factor is remarkably easy to tune, and the policy's abstention behavior is highly responsive to its adjustments. In conclusion, the dynamic reward mechanism of \texttt{AWA-RL} provides a vastly more robust and scalable pathway toward genuinely reliable search agents.

\section{Methodology}
\subsection{Task Formulation}
We formulate agentic-search-based QA as an interactive decision-making process. Given a query $q \in \mathcal{Q}$, the agent policy $\pi_\theta$ iteratively retrieve evidence. The process terminates with a final output $y \in \mathcal{A} \cup \{\bot\}$, where $\mathcal{A}$ is the space of valid answers and $\bot$ denotes the abstention or refusal action (e.g., outputting ``not enough information'' when evidence is insufficient).

\subsection{Outcome-Reward RL Preliminaries}
SOTA search agents are predominantly trained using outcome-reward RL, optionally preceded by cold-start SFT. During RL, the environment provides a sparse binary reward $R(y, y^*) \in \{0, 1\}$, indicating whether the final answer $y$ matches the ground truth $y^*$. We adopt Group Relative Policy Optimization (GRPO)~\citep{shao2024deepseekmath} to update $\pi_\theta$. For each $q$, GRPO samples a group of $G$ trajectories and optimizes the policy using the relative advantage $A_i$, computed by normalizing the outcome rewards within the group. 

\subsection{\texttt{AWA-RL}: Abstention-Aware RL}

The standard binary reward paradigm fundamentally fails to supervise the abstention action $\bot$. To address this and explicitly optimize the capability-reliability trade-off, \texttt{AWA-RL} reformulates the final environment reward $R(y, y^*, q_i)$ provided to the GRPO algorithm. We unify the reward structure by introducing a dynamically adjusted refusal reward $r_{ref}(q_i)$ specifically for abstention:
$$
R(y, y^*, q_i) = 
\begin{cases} 
1, \text{if } y = y^* \text{ \scriptsize (Correct)} \\
0, \text{if } y \neq y^* \text{ and } y \neq \bot \text{ \scriptsize (Incorrect)} \\
r_{ref}(q_i), \text{if } y = \bot \text{ \scriptsize (Abstention)}
\end{cases}
$$

To effectively calibrate the model's confidence and maintain intra-batch variance for stable RL updates, $r_{ref}(q_i)$ must account for both the intrinsic difficulty of the specific query and the real-time training dynamics. Therefore, we decompose $r_{ref}(q_i)$ into a query-aware base reward and a batch-level dynamic penalty:
$$
r_{ref}(q_i) = R_{base}(q_i) - P(E_t)
$$
where $R_{base}(q_i)$ establishes a refusal boundary based on the model's estimated capability, and $P(E_t)$ acts as a dynamic stabilizer to correct empirical over-or under-refusal during the RL rollout.

\paragraph{Estimating the Base Refusal Reward ($R_{base}$).}
Before abstention-aware RL training, we gauge the model's baseline capability by evaluating a standard, abstain-free checkpoint to compute the estimated correctness probability $p_i = P(y = y^* | q_i)$ for each training query $q_i$. To prevent premature abstention on inherently difficult questions, we apply a lower bound clip $p_{clip} = \max(p_i, \epsilon_1)$ (e.g., $\epsilon_1 = 0.05$). Furthermore, because executing a multi-step search is computationally and logically harder than simply generating a refusal token, models are naturally biased toward ``lazy refusal.'' To explicitly counteract this inertia and encourage search exploration, we apply a non-linear transformation $p_i^* = (p_{clip})^\gamma$. $\gamma \in (0, 1]$ is the ``courage'' factor, which scales the model's confidence. The query-specific base refusal reward is then defined as $R_{base}(q_i) = 1 - p_i^*$. This creates a theoretical equilibrium: for queries with high estimated capability, the expected reward of attempting an answer ($p_i^*$) mathematically exceeds that of refusing ($1 - p_i^*$), actively driving the agent to answer.

\paragraph{Computing the Dynamic Penalty ($P(E_t)$).}
During the RL phase, the empirical refusal rate naturally fluctuates and may deviate from the theoretical expectation. To compute the penalty $P(E_t)$, we first define the expected batch-level refusal rate as $\mu_{ref} = 1 - \frac{1}{B}\sum_{i=1}^B p_i^*$, where $B$ is the batch size. To avoid division by zero and unbounded penalties, we clip this expectation to $\mu_{clip} = \max(\mu_{ref}, \epsilon_2)$ (e.g., $\epsilon_2 = 0.05$). We then define the batch-level excess refusal rate $E_t$ at training step $t$ as:
$$
E_t = \frac{\hat{\mu}_{ref} - \mu_{clip}}{\mu_{clip}}
$$
where $\hat{\mu}_{ref} \in [0, 1]$ is the empirical mean refusal rate of the current rollout batch. Consequently, $E_t$ is strictly bounded within $[-1, E_{max}]$, where the upper bound $E_{max} = (1 - \mu_{clip}) / \mu_{clip}$.

To penalize policy collapse and guide the model back to optimal boundaries, we dynamically adjust the final refusal reward $r_{ref} \in [-1, 1]$ via a piecewise penalty function $P(E_t)$. This function is designed such that severe over-refusal ($E_t \to E_{max}$) pushes $r_{ref} \to -1$, while severe under-refusal ($E_t \to -1$) pushes $r_{ref} \to 1$:
$$
P(E_t) = 
\begin{cases} 
\frac{E_t}{E_{max}} \cdot (R_{base}(q_i) + 1), & \text{if } E_t \ge 0 \\
E_t \cdot (1 - R_{base}(q_i)), & \text{if } E_t < 0 
\end{cases}
$$
By subtracting this penalty from $R_{base}$, our framework ensures that the policy dynamically self-corrects while honoring the query-specific capability estimation.

In addition to reward shaping, we apply several training recipes to stabilize general RL training. More details can be found in Appendix \ref{appendix:training_recipes}.

\section{Experimental Setup}

\paragraph{Datasets \& Implementaion Details.}
We evaluate  on three challenging multi-hop QA benchmarks: HotpotQA~\citep{yang2018hotpotqa}, 2WikiMultiHopQA~\citep{ho2020constructing}, and MuSiQue~\citep{trivedi2022musique}. Our primary experiments utilize \texttt{Qwen2.5-7B-Instruct}~\citep{Yang2024Qwen25TR} as the base model. To establish the cold-start setting, the model is first fine-tuned on 5,000 search-augmented reasoning trajectories generated by a stronger teacher model (\texttt{GPT-oss-120b}\footnote{An open-weight model publicly available at \url{https://huggingface.co/openai/gpt-oss-120b}.}). This bootstraps the agent's initial multi-step search capabilities while preserving sufficient action-space entropy. 
Subsequently, we conduct reinforcement learning via GRPO on 5,000 filtered queries sampled from the training splits. The agent interacts with a standard Wikipedia search environment. During the RL phase, we employ \texttt{GPT-oss-120b} as the external LLM judge to automatically evaluate both final answer correctness and intermediate citation quality. More implementation details can be found in Appendix~\ref{appendix:exp_details}.

\paragraph{Evaluation Metrics.}
To comprehensively assess the capability-reliability trade-off, let $N_c$, $N_r$, and $N_w$ denote the number of correct, refused (abstained), and wrong responses, respectively, where $N_{total} = N_c + N_r + N_w$. Following the evaluation principles of \citet{xu2024rejection}, we measure policy performance using three metrics. \textbf{Accuracy (Acc)}, defined as $N_c / N_{total}$, evaluates the model's overall helpfulness and serves as our Recall. \textbf{Precision (Prec)}, computed as $N_c / (N_c + N_w)$, reflects the model's self-knowledge by measuring correctness strictly when it chooses to answer. Finally, we introduce \textbf{Reliability-Aware F1 (\texttt{RA-F1})}. Calculated as the harmonic mean of Precision and Accuracy ($\text{RA-F1} = 2 \cdot \text{Prec} \cdot \text{Acc} / (\text{Prec} + \text{Acc})$), this single metric holistically captures the optimal balance between answering confidently and abstaining safely.

\paragraph{Baselines.}
We compare \texttt{AWA-RL} against three representative abstention paradigms. \textbf{SFT-Mixed} injects 10\%--30\% teacher-generated refusal trajectories into the SFT data to test supervised elicitation. \textbf{RL-Unanswerable} mixes 10\%--30\% inherently unanswerable queries into the RL training set, optimizing via standard binary-reward GRPO. Finally, \textbf{Static Reward} applies RL with a fixed refusal reward $r_{ref}$ (e.g., $0.5$ or $0.9$), explicitly discouraging answers unless the model's confidence exceeds this static threshold.

\section{Experimental Results}
\begin{table*}[t] 
    \centering
    
    \begin{minipage}[t]{0.32\linewidth}
        \centering
        \caption{Performance with injected refusal or unanswerable data.}
        \label{tab:data_ratio_baselines}
        \resizebox{0.95\linewidth}{!}{%
            \begin{tabular}{lcccc}
                \toprule
                \textbf{Ratio} & \textbf{Acc} & \textbf{Refuse} & \textbf{Prec} & \textbf{\texttt{RA-F1}} \\
                \midrule
                \multicolumn{5}{c}{\small \textit{SFT with XX\% refusal trajectories}} \\
                0\%  & 0.530 & 0.000 & 0.530 & 0.530 \\
                10\% & 0.525 & 0.109 & 0.593 & 0.555 \\
                20\% & 0.512 & 0.131 & 0.595 & 0.548 \\
                30\% & 0.505 & 0.107 & 0.570 & 0.534 \\
                \midrule
                \multicolumn{5}{c}{\small \textit{RL from scratch with XX\% unanswerable questions}} \\
                0\%  & 0.472 & 0.005 & 0.474 & 0.473 \\
                10\% & 0.483 & 0.060 & 0.510 & 0.496 \\
                20\% & 0.425 & 0.293 & 0.578 & 0.487 \\
                30\% & 0.305 & 0.522 & 0.588 & 0.394 \\
                \midrule
                \multicolumn{5}{c}{\small \textit{Cold-start RL with XX\% unanswerable questions}} \\
                0\%  & 0.544 & 0.002 & 0.545 & 0.545 \\
                10\% & 0.545 & 0.001 & 0.545 & 0.545 \\
                20\% & 0.551 & 0.004 & 0.553 & 0.552 \\
                30\% & 0.557 & 0.018 & 0.567 & 0.562 \\
                \bottomrule
            \end{tabular}%
        }
    \end{minipage}\hfill 
    \begin{minipage}[t]{0.32\linewidth}
        \centering
        \caption{Performance under a static refusal reward ($r_{ref}$).}
        \label{tab:static_reward}
        \resizebox{1\linewidth}{!}{%
            \begin{tabular}{lcccc}
                \toprule
                \boldmath{$r_{ref}$} & \textbf{Acc} & \textbf{Refuse} & \textbf{Prec} & \textbf{\texttt{RA-F1}} \\
                \midrule
                \multicolumn{5}{c}{\small \textit{RL from scratch with Static Refusal Reward}} \\
                0.00 & 0.493 & 0.017 & 0.500 & 0.496 \\
                0.05 & 0.000 & 0.999 & 0.000 & 0.000 \\
                0.10 & 0.000 & 0.999 & 0.000 & 0.000 \\
                0.20 & 0.000 & 0.921 & 0.000 & 0.000 \\
                \midrule
                \multicolumn{5}{c}{\small \textit{Cold-start RL with Static Refusal Reward}} \\
                0.00 & 0.552 & 0.012 & 0.559 & 0.556 \\
                0.20 & 0.488 & 0.239 & 0.649 & 0.553 \\
                0.40 & 0.469 & 0.338 & 0.708 & 0.561 \\
                0.60 & 0.438 & 0.421 & 0.750 & 0.550 \\
                0.80 & 0.426 & 0.441 & 0.747 & 0.540 \\
                \bottomrule
            \end{tabular}%
        }
    \end{minipage}\hfill
    \begin{minipage}[t]{0.32\linewidth}
        \centering
        \caption{Performance of \texttt{AWA-RL} under varying ``courage'' factor.}
        \label{tab:awa_rl_results}
        \resizebox{0.95\linewidth}{!}{%
            \begin{tabular}{lcccc}
                \toprule
                \boldmath{$\gamma$} & \textbf{Acc} & \textbf{Refuse} & \textbf{Prec} & \textbf{\texttt{RA-F1}} \\
                \midrule
                \multicolumn{5}{c}{\small \textit{\texttt{AWA-RL} from scratch with varying encouragement}} \\
                1.00  & 0.322 & 0.360 & 0.475 & 0.381 \\
                0.50  & 0.381 & 0.395 & 0.605 & 0.459 \\
                0.25  & 0.441 & 0.269 & 0.581 & 0.499 \\
                0.20  & 0.459 & 0.246 & 0.591 & \textbf{0.515} \\
                0.125 & 0.487 & 0.104 & 0.537 & 0.511 \\
                0.00  & 0.493 & 0.017 & 0.500 & 0.496 \\
                \midrule
                \multicolumn{5}{c}{\small \textit{Cold-start \texttt{AWA-RL} with varying encouragement}} \\
                1.00  & 0.458 & 0.383 & 0.738 & 0.561 \\
                0.50  & 0.484 & 0.306 & 0.699 & 0.568 \\
                0.25  & 0.509 & 0.225 & 0.664 & 0.573 \\
                0.20  & 0.529 & 0.192 & 0.662 & \textbf{0.585} \\
                0.125 & 0.545 & 0.125 & 0.628 & 0.582 \\
                0.00  & 0.552 & 0.012 & 0.559 & 0.556 \\
                \bottomrule
            \end{tabular}%
        }
    \end{minipage}
\end{table*}
\begin{figure*}[t]
    \centering
    \includegraphics[width=0.9\textwidth]{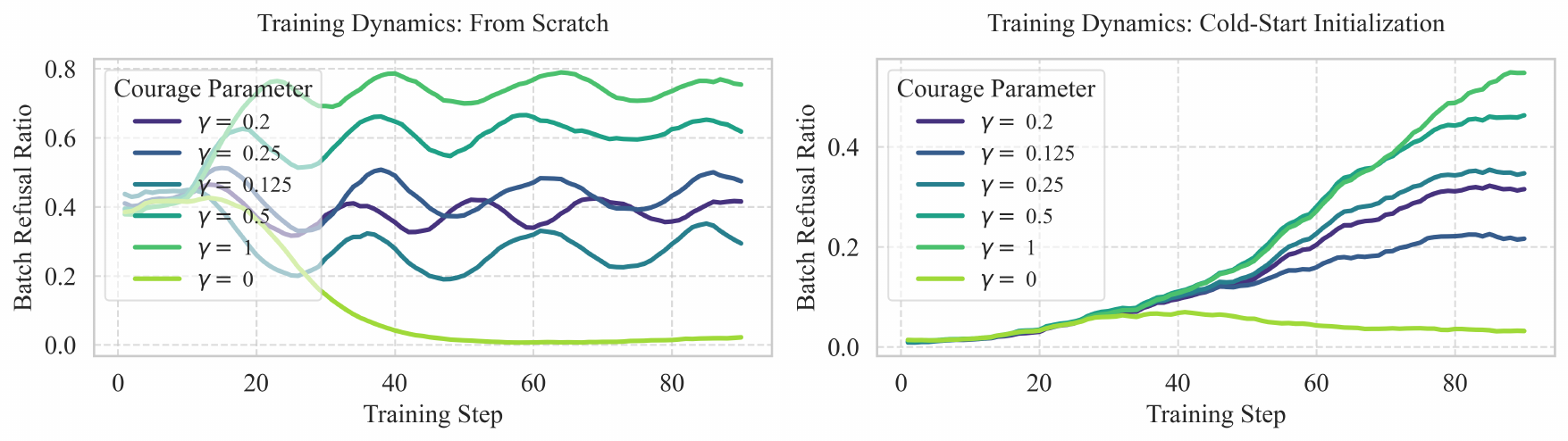}
    \caption{Training dynamics of batch refusal ratios over 90 GRPO steps under different $\gamma$ settings. }
    \label{fig:training_dynamics}
\end{figure*}

In this section, we evaluate the capability-reliability trade-off. Due to space constraints, averaged metrics across our three benchmarks are reported here (see Appendix~\ref{appendix:full_results} and \ref{appendix:rollout_cases} for per-dataset breakdowns and rollout cases).

\paragraph{Failures of Conventional Baselines.}
Tables \ref{tab:data_ratio_baselines} and \ref{tab:static_reward} demonstrate that static training paradigms fail to establish a reliable abstention boundary. First, manually manipulating data proportions yields highly unpredictable behaviors. SFT alone struggles to internalize abstention boundaries (refusal rates stagnate around 10-13\%). During RL, data manipulation triggers either a ``lazy refusal'' collapse (e.g., 52.2\% refusal from scratch) or stubborn overconfidence (ignoring unanswerable queries in cold-start). 
Second, applying a static refusal reward ($r_{ref}$) acts as a blunt instrument. In from-scratch RL, even a marginal $r_{ref}=0.05$ causes catastrophic reward hacking (99.9\% refusal). In cold-start settings, scaling $r_{ref}$ forces a rigid zero-sum trade-off: Precision improves at the severe expense of Accuracy (Recall).

\paragraph{\texttt{AWA-RL} Achieves Optimal and Stable Balance.}
In contrast, \texttt{AWA-RL} cleanly resolves these limitations by dynamically anchoring rewards to query-specific prior capabilities (Table \ref{tab:awa_rl_results}). The ``courage'' factor $\gamma$ serves as a highly responsive dial for confidence mapping. At the optimal ``sweet spot'' ($\gamma = 0.2$), \texttt{AWA-RL} achieves optimal balances in both RL settings. Especially for the cold-started policy, \texttt{AWA-RL} hits a state-of-the-art \texttt{RA-F1} of 0.585, boosting Precision from 0.559 to 0.662 with only a marginal drop in Accuracy (0.552 to 0.529), breaking the zero-sum constraints of static rewards. Furthermore, \texttt{AWA-RL} yields remarkably stable training dynamics (Figure \ref{fig:training_dynamics}). Instead of wild oscillations or rapid reward hacking, the batch-level refusal rates smoothly converge to distinct, predictable plateaus within 20-30 steps corresponding to $\gamma$. This confirms that our dynamic reward formulation provides a robust, scalable, and controllable pathway to reliably self-aware agents.

\section{Conclusion}
We introduced \texttt{AWA-RL} to resolve the critical capability-reliability trade-off in LLM agents. While conventional methods, such as data manipulation or static rewards, inevitably trigger stubborn overconfidence or catastrophic reward hacking, \texttt{AWA-RL} dynamically anchors the refusal reward to the model's query-specific prior capabilities. Controlled by a single, highly responsive ``courage'' factor ($\gamma$), our framework successfully decouples capability acquisition from abstention calibration. By maximizing overall system utility (\texttt{RA-F1}), \texttt{AWA-RL} provides a robust and scalable pathway toward genuinely reliable search agents.

\newpage

\section*{Limitations}
While \texttt{AWA-RL} demonstrates significant improvements in optimizing the capability-reliability trade-off, we acknowledge a few limitations that pave the way for future research.

\paragraph{Scale and Task Modality.} 
Our current empirical evaluations are primarily focused on the \texttt{Qwen2.5-7B-Instruct} model and multi-hop agentic search QA tasks. Although the theoretical formulation of our dynamic reward shaping is inherently model-agnostic and broadly applicable to general agentic RL domains (e.g., complex tool-use, mathematical reasoning, and coding), scaling the validation to larger foundational models and a wider spectrum of task modalities remains an important next step to confirm its universal efficacy.

\paragraph{Dependence on the Courage Factor.} 
\texttt{AWA-RL} successfully distills the complex balance between answering and abstaining into a single, intuitive hyperparameter: the courage factor $\gamma$. As demonstrated in our training dynamics, the policy is highly responsive to $\gamma$, making it practically straightforward to tune. Nevertheless, identifying the strictly optimal value still requires empirical adjustments based on specific dataset difficulties and model initializations. Our future research will explore fully end-to-end, adaptive mechanisms that can auto-calibrate this confidence mapping without human intervention.

\paragraph{Computational Overhead for Prior Estimation.} 
Establishing the optimal base refusal reward ($R_{base}$) requires a preliminary inference pass over the training set to calculate the prior success rate ($p_i$). While this offline sampling introduces an initial computational cost, we argue it is a highly worthwhile trade-off: it provides the most faithful and unbiased estimation of the model's intrinsic capabilities, which is the cornerstone for precise refusal boundary control during RL. Still, developing sample-free or computationally lighter proxy metrics to estimate model capability presents an exciting avenue for subsequent optimization.

\section*{Ethical Considerations}

Our work primarily contributes to the safety and reliability of Large Language Models by explicitly training them to abstain from answering when sufficient evidence is lacking. By addressing the capability-reliability trade-off in agentic QA systems, our \texttt{AWA-RL} framework directly mitigates the risks of hallucination and overconfident factual errors, which is a crucial step toward deploying trustworthy AI in real-world scenarios.

Despite these positive impacts, several ethical implications warrant careful consideration:

\paragraph{Disparities in Abstention.} 
In our framework, the abstention boundary is intrinsically anchored to the model's prior capability estimation ($p_i$). It is well-documented that pre-trained language models often exhibit performance disparities, generally performing worse on topics related to underrepresented demographics, minority cultures, or low-resource languages. Consequently, our dynamic reward shaping might disproportionately encourage the agent to refuse queries from these marginalized domains. While abstaining is safer than hallucinating, this behavior could inadvertently exacerbate inequalities in information access, as users querying non-mainstream topics might consistently receive ``not enough information'' responses.

\paragraph{Bias in External Knowledge and Evaluators.} 
Our agent's search and reasoning capabilities are optimized interacting with an external corpus (e.g., Wikipedia) and supervised by an LLM-based judge for citation rewards. Any inherent biases, historical prejudices, or misinformation embedded within the retrieval environment can shape the agent's definition of ``truth.'' Furthermore, LLM judges may harbor implicit preferences for certain linguistic styles or mainstream viewpoints. If the judge incorrectly penalizes valid but minority perspectives, the policy might learn to improperly reject valid queries or confidently output biased information.

Future deployments of abstention-aware agents should pair our dynamic reward mechanisms with rigorous bias audits, ensuring equitable capability expansion across diverse user groups and topics.

\bibliography{custom}

\newpage

\appendix
\section*{Appendix}

\startcontents[sections]
\printcontents[sections]{l}{1}{\setcounter{tocdepth}{2}}

\section{Related Work}
\label{appendix:related_work}
\paragraph{Agentic Search for Open-domain QA.}
Recent progress in open-domain QA has increasingly shifted from one-shot retrieval-augmented generation to \emph{agentic search}, where LLMs interleave reasoning with explicit actions such as query reformulation, retrieval, and evidence integration. Early paradigms such as ReAct~\citep{yao2022react} and Toolformer~\citep{schick2023toolformer} established the value of coupling reasoning with external actions and tool use, showing that iterative interaction with external knowledge sources can substantially improve factuality and task performance. Building on this line, more recent search and deep research agents train models to perform multi-turn retrieval and synthesis in a more end-to-end manner. For example, Search-R1~\citep{jin2025search} optimizes interleaved search-and-reason trajectories with reinforcement learning and a simple outcome-based reward, while recent deep research systems combine iterative web search with preference tuning, supervised fine-tuning, or RL to improve long-horizon reasoning and report generation \citep{ yamada2025open, hu2025step, fan2025understanding, zhang20252}. These advances have significantly strengthened the capability frontier of search-based QA agents.

However, existing agentic search training paradigms are still largely \emph{correctness-centric}. In representative RL formulations such as Search-R1, the reward is defined only over the final outcome correctness, with no explicit incentive for refusing unsupported answers~\citep{jin2025search, team2025tongyi, team2025mirothinker, chu2026redsearcher, tao2025webshaper, feng2025retool}. Likewise, recent deep research systems primarily optimize answer quality, factuality, or task success, rather than calibrating when the agent should abstain. As a result, once retrieval fails, evidence is insufficient, or the problem lies beyond the agent's effective capability boundary, the policy is still biased toward producing an answer. Our work addresses precisely this missing dimension: instead of treating abstention as a post-hoc patch, we incorporate it directly into the same RL process that trains search behavior, so that capability improvement and reliability calibration are optimized jointly rather than separately.

\paragraph{Abstention Training.}
A growing body of work studies how to make LLMs refuse questions that fall outside their knowledge boundary. One line of research formulates refusal as a reliability-alignment problem. RLKF~\citep{xu2024rejection} trains a reward model from knowledge feedback so that refusal is preferred on out-of-knowledge questions, while EKBM~\citep{zheng2025enhancing} explicitly models knowledge boundaries through a fast-slow framework that refines uncertain cases with an additional reasoning stage. Another line focuses on the adverse effect of capability-oriented post-training:~\citet{song2025hallucination} show that standard reinforcement finetuning can sharply reduce refusal behavior and increase hallucination on synthetic unanswerable math problems, and~\citet{kirichenko2025abstentionbench} further show that modern reasoning models often exhibit degraded abstention despite higher reasoning accuracy. More task-specific approaches include training abstention on temporal QA with CoT supervision and abstention-aware RL~\citep{zhou2026silence}, as well as reliability alignment for tool use by expanding the action space with indecisive actions and optimizing on synthesized data~\citep{xu2024reducing}.

Despite their promise, these methods leave several open issues for search agents. First, most prior abstention work is developed in \emph{single-turn} or otherwise non-agentic settings, such as math QA, temporal QA, dialogue state tracking, or tool selection, where the model's capability boundary is comparatively static~\citep{xu2024rejection, song2025hallucination, zhou2026silence, zheng2025enhancing, xu2024reducing}. In contrast, for search agents, the answerability of a question changes dynamically as the agent interacts with the environment and accumulates evidence. Second, many approaches rely on synthesized unanswerable data, abstained variants of existing benchmarks, or auxiliary reward models/modules, which may introduce task-specific heuristics and reduce portability across models with different intrinsic capabilities~\citep{song2025hallucination, zhou2026silence, xu2024rejection, xu2024reducing}. Third, existing reward designs often treat abstention with a fixed bonus or through a separate post-hoc stage, which makes it difficult to preserve the model's willingness to answer when sufficient evidence is actually available~\citep{zhou2026silence, zheng2025enhancing}. Our proposed \texttt{AWA-RL} is designed to fill this gap: it learns abstention \emph{within} search-agent RL, uses a dynamically shaped refusal reward tied to an estimated prior capability rather than a static refusal bonus, and adds an explicit generation-encouragement term to avoid collapse into lazy refusal. In this way, our method targets the core unresolved problem left by prior work: how to jointly optimize search capability and abstention reliability under a unified, lightweight RL objective.

\section{Additional Recipes for Stable RL}
\label{appendix:training_recipes}

In this section, we provide detailed descriptions of the auxiliary training mechanisms utilized to stabilize the reinforcement learning process and to continuously enhance the underlying search capabilities of the agent.

\subsection{Reward Stabilization via EMA Smoothing and Step-wise Updates}
During the on-policy rollout phase, the empirical refusal rate ($\hat{\mu}_{ref}$) can fluctuate significantly from step to step simply due to the random sampling of prompts and the inherent stochasticity of the model's generation. If the dynamic penalty $P(E_t)$ is updated immediately based on these noisy step-wise fluctuations, the reward landscape will become overly jittery, leading to unstable policy updates and preventing the model from converging to an optimal abstention boundary.

To mitigate this variance, we introduce two robust stabilization mechanisms:
\begin{enumerate}
    \item \textbf{Exponential Moving Average (EMA) Smoothing:} Instead of using the instantaneous excess refusal rate $E_t$, we maintain a smoothed running average $\bar{E}_t$ across training steps:
    $$
    \bar{E}_t = \alpha \bar{E}_{t-1} + (1 - \alpha) E_t
    $$
    where $\alpha \in [0, 1)$ is the decay factor (set to $\alpha = 0.5$ in our experiments). This EMA acts as a low-pass filter, ensuring that the penalty only responds to sustained trends in over-refusal or under-refusal rather than transient batch noise.
    
    \item \textbf{Step-wise Quantization:} To further prevent microscopic, continuous shifts in the refusal reward, we enforce discrete, step-wise updates by quantizing the smoothed excess rate $\bar{E}_t$. Given a predefined step-tier threshold $\tau$ (e.g., $\tau = 0.1$), the effective excess rate is discretized as:
    $$
    \tilde{E}_t = \tau \lfloor \bar{E}_t / \tau \rceil
    $$
    where $\lfloor \cdot \rceil$ denotes rounding to the nearest integer.
\end{enumerate}

Ultimately, the dynamic penalty is calculated as $P(\tilde{E}_t)$ rather than $P(E_t)$. This quantization creates a ``staircase'' effect: the final refusal reward $r_{ref}$ remains strictly constant unless the model's smoothed refusal behavior crosses a significant threshold, thereby providing a much more stable optimization target for the GRPO algorithm.

\subsection{Dense Citation Reward and Composite Objective}
Relying exclusively on the sparse outcome reward, which assigns $1$ for correct, $0$ for incorrect, and $r_{ref}$ for abstention, can be highly inefficient for training complex agentic tasks. When the model receives a $0$, it lacks granular feedback on \textit{why} it failed (e.g., did it fail to retrieve the right documents, or did it fail to reason over them?).

To provide a dense learning signal that continuously bolsters the model's multi-hop search and reasoning skills, we introduce an LLM-judged citation reward. During generation, the agent is prompted to cite specific supporting document IDs for its claims. An external, powerful LLM acts as a judge to evaluate these citations across two dimensions:
\begin{itemize}
    \item \textbf{Citation Coverage ($R_{cov}$):} The judge evaluates whether the cited documents contain all the necessary logical hops to derive the final answer, outputting a continuous score $R_{cov} \in [0, 1]$.
    \item \textbf{Citation Redundancy ($R_{red}$):} To prevent ``citation hacking'' (where the agent trivially cites every retrieved document to maximize coverage), the judge simultaneously checks for irrelevant citations, assigning a discrete penalty score $R_{red} \in \{0, 0.25, 0.5, 1\}$.
\end{itemize}

For the final GRPO update, we combine these dense signals with our sparse outcome reward. To prevent rewards of different scales from dominating the gradient, we independently whiten (z-score normalize) each component within the sampled generation group. Let $\text{Wh}(\cdot)$ denote this intra-group whitening operation. The final composite advantage $A_i$ for a specific trajectory is formulated as a weighted sum:
$$
\scriptsize A_i = 2 \cdot \text{Wh}(R(y, y^*, q_i)) + 3 \cdot \text{Wh}(R_{cov}) - 1 \cdot \text{Wh}(R_{red})
$$
This composite objective represents a highly effective decoupling strategy. By heavily weighting the dense citation metrics (ratio of 3 for coverage), the agent steadily optimizes its fine-grained search and grounding skills. Meanwhile, the dynamically scaled outcome reward (ratio of 2) is freed up to be exclusively dedicated to balancing the overarching capability-reliability (answer-versus-abstain) trade-off.

\section{Detailed Experimental Setup}
\label{appendix:exp_details}

In this section, we provide the comprehensive implementation details, environment configurations, and hyperparameters omitted from the main text.

\subsection{Datasets and Search Environment}
\paragraph{Evaluation and Training Data.} 
For the main evaluation, we randomly sample 1,000 queries from the official validation sets of HotpotQA, 2WikiMultiHopQA (2Wiki), and MuSiQue. For the RL training phase, we utilize the training splits of 2Wiki and MuSiQue. To ensure the efficiency of the GRPO updates, we conduct a preliminary test rollout on the training set to filter out queries that are either trivially easy (where the model answers correctly in all sampled trajectories) or impossibly hard (where the model fails in all trajectories). Such queries provide zero relative advantage and contribute little to policy learning. From the remaining filtered pool, we randomly sample 5,000 queries for the final RL training.

\paragraph{Search Environment.}
We adopt a search environment configuration identical to Search-R1. Specifically, we use the 2018 Wikipedia dump, partitioned into 100-word chunks. These chunks are embedded using the E5 model~\citep{wang2022text} and indexed using FAISS~\citep{douze2024faiss} for efficient dense retrieval. At each query step, the retriever returns the top-5 passages based on embedding cosine similarity.

\subsection{Implementation and Hyperparameters}
\paragraph{Model Initialization.}
All primary experiments are conducted using the \texttt{Qwen2.5-7B-Instruct} model, which inherently possesses strong foundational potential for tool use. For our cold-start setting, we generate 5,000 distinct, successful trajectories using a highly capable teacher model (\texttt{GPT-oss-120b}). The base Qwen model undergoes Supervised Fine-Tuning (SFT) on this dataset for 2 epochs, successfully bootstrapping its search and reasoning skills before RL intervention.

\paragraph{RL Configuration.}
During the reinforcement learning phase, we train the model for 5 epochs using the GRPO algorithm. The global batch size is set to 256, the PPO mini-batch size is 256, and the group rollout size ($G$) is set to 16. We use a learning rate of $1 \times 10^{-6}$. The maximum context length is capped at 16,384 tokens, and the agent is allowed a maximum of 20 interaction turns with the search environment per query.

\paragraph{Entropy Control and Judges.}
To maintain exploration and prevent premature policy collapse during RL, we completely remove the standard KL divergence penalty in GRPO. Instead, we apply the dynamic entropy control mechanism introduced in Skywork-OR1~\citep{he2025skywork}, strictly keeping the policy entropy above a threshold of 0.4. Furthermore, two separate \texttt{GPT-oss-120b} instances are employed as LLM judges: one dedicated to evaluating the rigorous correctness of the final answer, and the other to assessing citation quality (coverage and redundancy). All prompt templates used for the agent and the judges are provided in Appendix~\ref{appendix:prompt_templates}. 

\paragraph{Hardware.}
All training and evaluation pipelines are executed on two compute nodes, each equipped with 8$\times$NVIDIA H100 GPUs.

\section{Prompt Templates}
\label{appendix:prompt_templates}
\subsection{Prompt for RL Training}
\label{appendix:rl_sys_prompt}
\begin{mdbox}[label={box:rl_sys_prompt}]{Prompt for RL Training}
\begin{mdverbatim}
You are an assistant tasked with answering user queries step-by-step using the `wikipedia_search` tool.

Each use of `wikipedia_search` returns a list of documents. Each document includes:

* `id`: a unique identifier in the format `"x-y"`, where `x` indicates the search iteration number and `y` indicates the index within that result set (e.g., `"1-1"`, `"1-2"`, `"2-1"`).
* `content`: the title and a chunk of text from the Wikipedia article.

Tool responses will be returned in the following format:

<tool_response>
```json
[
  {
    "doc_id": "x-y",
    "content": "\"Document Title\"\nText content of the document..."
  },
  ...
]
```
</tool_response>

Your task is to explore the question systematically and provide reasoning and answer with precise citations from the returned documents.

### Workflow

* Before calling the tool, always generate a `<think>` block explaining your reasoning.
* For the final answer, use an `<answer>` block containing a JSON object with:
  * `answer`: a short answer to the question.
  * `rationale`: a concise explanation of how the sources support the answer.
  * `citations`: a list of doc_ids
* You can also return a refusal ("not enough information") if the question cannot be answered based on the search engine.

### Required Output Format

#### Reasoning Block

Enclose your reasoning within `<think> </think>` tags, for example:

<think>  
Your reasoning process here...  
</think>

#### Final Answer Block

Enclose your structured answer within `<answer> </answer>` tags, for example:

<answer>
```json
{
  "answer": "A short answer to the question, preferably in one or few words. Or 'not enough information' if the question cannot be answered.",
  "rationale": "Concise justification of the logic that leads to the answer.",
  "citations": ["x-y", "a-b", ...]
}
```
</answer>

# Tools

You may call one or more functions to assist with the user query.

You are provided with function signatures within <tools></tools> XML tags:
<tools>
{
  "type": "function",
  "function": {
    "name": "wikipedia_search",
    "description": "Search Wikipedia for information about a specific query. Returns a list of summaries of the related articles.",
    "parameters": {
      "type": "object",
      "properties": {
        "query": {
          "type": "string",
          "description": "The specific query term to search on Wikipedia."
        },
        "top_n": {
          "type": "integer",
          "description": "The number of documents to return."
        }
      },
      "required": ["query", "top_n"]
    }
  }
}
</tools>

For each function call, return a json object with function name and arguments within <tool_call></tool_call> XML tags:
<tool_call>
{"name": <function-name>, "arguments": <args-json-object>}
</tool_call>
\end{mdverbatim}
\end{mdbox}

\subsection{Prompt for LMJudge}
\label{appendix:prompt_lmjudge}
\begin{mdbox}[label={box:prompt_lmjudge}]{Prompt for LMJudge}
\begin{mdverbatim}
You will be given a question and its ground truth answer list where each item can be a ground truth answer. Provided a pred_answer, you need to judge if the pred_answer correctly answers the question based on the ground truth answer list. You should first give your rationale for the judgement, and then give your judgement result (i.e., correct or incorrect).

Here is the criteria for the judgement:
1. The pred_answer doesn't need to be exactly the same as any of the ground truth answers, but should be semantically same for the question.
2. Each item in the ground truth answer list can be viewed as a ground truth answer for the question, and the pred_answer should be semantically same to at least one of them.

question: {question}
ground truth answers: {gt_answer}
pred_answer: {pred_answer}

The output should in the following json format:
```json 
{{
    "rationale": "your rationale for the judgement, as a text",
    "judgement": "your judgement result, can only be 'correct' or 'incorrect'"
}}
```

Your output:
\end{mdverbatim}
\end{mdbox}

\subsection{Prompt for Citation Reward}
\label{appendix:prompt_citation_reward}
\begin{mdbox}[label={box:prompt_citation_reward}]{Prompt for Citation Reward}
\begin{mdverbatim}
## Role
You are an expert Citation Quality and Coverage Evaluator.

## Input
You will be provided with:
1. **Question**: The user query.
2. **Ground Truth (GT) Answer**: The correct answer string.
3. **Ground Truth (GT) Atomic Evidence**: A list of specific supporting facts that should be cited. Each atomic evidence represents one key piece of information.
4. **Model Citations**: The citations/evidence passages that the model has provided in its answer.

## Goal
Evaluate the model's citations on TWO dimensions:
A) **Coverage Count**: How many GT Atomic Evidence items are covered by the citations?
B) **Redundancy/Irrelevance Severity**: The severity level of redundant or irrelevant citations (0-3 scale).

## Evaluation Procedure

### A) Coverage Count

**Task**: Count how many items from the GT Atomic Evidence are adequately covered in the Model Citations.

**Counting Rules**:
1. Go through EACH item in the GT Atomic Evidence list one by one.
2. For each GT item, check if the Model Citations contain the **same semantic information**:
   - Exact wording is NOT required
   - BUT specific details (names, dates, numbers, entities) MUST match
   - Vague references do NOT count as coverage
3. If a GT item is sufficiently covered (even partially across multiple citations), count it as **1**.
4. If a GT item is completely missing or only vaguely referenced, count it as **0**.

**Output**: `covered_count` = integer, the total number of GT Atomic Evidence items that are covered.

**Example**:
- GT Atomic Evidence has 5 items: [A, B, C, D, E]
- Model Citations cover: A (fully), B (fully), C (missing), D (vaguely), E (fully)
- covered_count = 3 (only A, B, E count)

### B) Redundancy/Irrelevance Severity (0-3 scale)

**Task**: Assess the severity of redundant or irrelevant content in the Model Citations.

**Definitions**:
- **Redundant Citation**: A citation that repeats information already provided by another citation. The repeated information must be substantially the same fact.
- **Irrelevant Citation**: A citation that provides information NOT needed to support any GT Atomic Evidence or answer the question.

**Severity Levels**:

**0 (Clean - No Issues)**
- ALL citations are unique and relevant.
- Every citation contributes distinct, necessary information.
- No redundancy, no irrelevance.

**1 (Minor Issues)**
- A small portion of citations are redundant or irrelevant.
- 1-2 problematic citations, OR slight overlap between citations.
- Roughly <25% of citations are problematic.

**2 (Moderate Issues)**
- A noticeable portion of citations are redundant or irrelevant.
- Multiple problematic citations (3-5), OR significant redundancy where key facts are repeated.
- Roughly 25-50% of citations are problematic.

**3 (Severe Issues)**
- A large portion or majority of citations are redundant or irrelevant.
- Many problematic citations (5+), OR heavy redundancy with massive duplication.
- Roughly >50% of citations are problematic.

**Evaluation Guide**:
- Count total Model Citations: N
- Count redundant citations: R
- Count irrelevant citations: I
- Calculate proportion: (R + I) / N
- Map to severity: <25% → 1, 25-50% → 2, >50% → 3, 0% → 0

## Important Notes
- For Coverage: Only count a GT item as covered if the specific information is clearly present.
- For Severity: Base your judgment on the proportion of problematic citations.
- If Model Citations is empty/null, set covered_count=0 and redundancy_severity=0.

## Output Format
You must output your answer in a valid **JSON object**. Do not output any other text or explanations outside the JSON.

**JSON Schema:**
```json
{
  "covered_count": <integer>,
  "redundancy_severity": 0 | 1 | 2 | 3
}
```
\end{mdverbatim}
\end{mdbox}

\newpage

\section{\texttt{AWA-RL} Rollout Cases}
\label{appendix:rollout_cases}
\subsection{Correctly Answered}
\label{appendix:correct_case}

The following case demonstrates how the \texttt{AWA-RL} trained model performs step-by-step planning, reasoning, searching, and verification to arrive at the correct answer for a complex multi-hop question from the MuSiQue validation set.

\begin{casebox}[label={box:correct_case_1}]{\textbf{What happened to the Curonians who lived in the area where Erich Zakowski was born?} \\ \textbf{Reference Answer:} expelled by the Soviet Union}
\begin{caseverbatim}
<think>We need to answer: "What happened to the Curonians who lived in the area where Erich Zakowski was born?" Need to find Erich Zakowski birthplace, then find Curonians in that area, what happened to them. Likely about the Curonians (an ancient Baltic tribe) living in the region of Lithuania/Poland? Need to see if Erich Zakowski born in Curonian territory historically. Let's search.</think>

<tool_call>
{'name': 'wikipedia_search', 'arguments': {'query': 'Erich Zakowski birthplace', 'top_n': 5}}
</tool_call><|im_end|>
<|im_start|>user
<tool_response>[
  {
    "doc_id": "1-1",
    "content": "\"Erich Zakowski\"\nErich Zakowski Erich Zakowski (born 25 November 1934 in East Prussia) is a German master mechanic, and the founder and longtime head of the Zakspeed racing team. After the Second World War, Zakowski fled from Prussia with his mother and four siblings initially to Dortmund then Hamburg and finally the family settled in Niederzissen. Zakowski graduated in Andernach as an apprentice auto mechanic, and passed his master exam. He founded his own garage in Niederzissen, which was the location of the Zakspeed racing operation, starting in 1968. In 1968, under the name \"\"Zakowski Niederzissen tuning\"\", he started using a Ford"
  },
  {
    "doc_id": "1-2",
    "content": "\"Erich Zakowski\"\nEscort for the Eifel race on the Nürburgring. In the 1970s and 1980s, under the direction of Erich Zakowski, Zakspeed established itself in various racing series, (especially in sports car racing). Zakspeed entered the Formula 1 series in 1985. In 1990, after five years of racing, Zakowski retired active leadership of the team and handed it to his son Peter. Erich Zakowski Erich Zakowski (born 25 November 1934 in East Prussia) is a German master mechanic, and the founder and longtime head of the Zakspeed racing team. After the Second World War, Zakowski fled from Prussia with his mother and"
  },
  {
    "doc_id": "1-3",
    "content": "\"Henry Jarecki\"\nHenry Jarecki Dr. Henry George Jarecki (born April 15, 1933) is a German American academic, psychiatrist, entrepreneur, producer and philanthropist. Henry Jarecki was born into a Jewish family in Stettin, Germany, (now Szczecin, Poland), the son of Max Jarecki, a physician, and Gerda Kunstmann, the scion of a shipping family. As a child, he fled Nazi Germany with his family for the United Kingdom and subsequently the United States.His wealthy family was able to transfer their wealth from Germany. Jarecki graduated from the Medical Faculty at Heidelberg University in 1957, and subsequently spent more than a decade as an academic,"
  },
  {
    "doc_id": "1-4",
    "content": "\"Zdzisław Zakrzewski\"\nZdzisław Zakrzewski Zdzisław Zakrzewski, of the Jastrzębiec coat of arms (11 November 1919 in Lwów – 21 March 2013 in Hayward, California) was a Polish-American optical engineer, philanthropist, banker, a veteran of Poland's defensive struggle in September 1939, and a social and political activist. Zdzisław Zakrzewski graduated from the Classical Gymnasium of St. Jadwiga and studied engineering at the Lwów Polytechnic Institute. He was an altar boy, a boy scout, an activist of the Lwów chapter of the All-Polish Youth and the \"\"Brotherly Aid\"\" Association, as well as a member of the \"\"Ikaria\"\" Student Fraternity. In September 1939, after the"
  },
  {
    "doc_id": "1-5",
    "content": "\"Erich Reich\"\nErich Reich Sir Erich Arieh Reich (born 1935, Vienna, Austria) is an Austrian-born entrepreneur based in London, who through his charity challenges company Classic Tours has inspired over 42,000 people to raise £60million for 300+ UK charities. Reich was born in 1935. In 1938, he was one of the children of 5,000 families deported by Nazi Germany to Poland. Under the Kindertransport agreement, he arrived in the United Kingdom, aged 4, in August 1939, one of 10,000 children. He never saw his parents again, who were murdered in Auschwitz. Initially placed with a foster family in Dorking, Surrey, he moved"
  }
]</tool_response><|im_end|>
<|im_start|>assistant
<think>Erich Zakowski born in East Prussia. East Prussia historically includes Curonian territory. The Curonians were an ancient Baltic tribe living in the area of present-day Kaliningrad Oblast, Russia, and also parts of Lithuania. What happened to them? They were assimilated or disappeared; the region became German, then after WWII, the Germans expelled, area became Soviet, Curonian ethnic group extinct. Need source about Curonians and what happened to them. Let's search.</think>

<tool_call>
{'name': 'wikipedia_search', 'arguments': {'query': 'Curonians fate', 'top_n': 5}}
</tool_call><|im_end|>
<|im_start|>user
<tool_response>[
  {
    "doc_id": "2-1",
    "content": "Curonians\nCuronians The Curonians or Kurs (Curonian: \"\"Kursi\"\"; ; ; ; ; ; ; ) were a Baltic tribe living on the shores of the Baltic Sea in what are now the western parts of Latvia and Lithuania from the 5th to the 16th centuries, when they merged with other Baltic tribes. They gave their name to the region of Courland (\"\"Kurzeme\"\"), and they spoke the Old Curonian language. Curonian lands were conquered by the Livonian Order in 1266 and they eventually merged with other Baltic tribes participating in the ethnogenesis of Lithuanians and Latvians. Direct descendants of the Curonians include"
  },
  {
    "doc_id": "2-2",
    "content": "Curonians\npart of Samogitian ethnic self-identification. On the Latvian side during the Livonian War, the descendants of the Curonian nobility, although downgraded to peasant status, fought the Russians, as Johann Renner's chronicle reports: The Curonian language became extinct by the 16th century. Bishop Rimbert of Bremen (lived before 888 AD) in his life of St. Ansgar, Vita Ansgarii described the territory inhabited by the Curonians (\"\"Cori\"\") and gave the names of the administrative districts or lands (\"\"civitates\"\"): Curonians The Curonians or Kurs (Curonian: \"\"Kursi\"\"; ; ; ; ; ; ; ) were a Baltic tribe living on the shores of the"
  },
  {
    "doc_id": "2-3",
    "content": "\"Curonian Lagoon\"\n(); the southern part of the Spit and the Lagoon remained in Germany until 1945. This border is now the border between Lithuania and Russia, as after World War II, the southern end of the Spit and the German area south of the river — the part of East Prussia with the city Königsberg in Sambia — became part an exclave of Russia called Kaliningrad Oblast. While today the Kursenieki, also known as Kuršininkai, are a nearly extinct Baltic ethnic group living along the Curonian Spit, in 1649 Kuršininkai settlement spanned from Memel (Klaipėda) to Danzig (Gdańsk). The Kuršininkai were"
  },
  {
    "doc_id": "2-4",
    "content": "Curonians\nthe Kursenieki of the Curonian Spit and the so-called Curonian Kings of Courland. The ethnic origin of the Curonians has been disputed in the past. Until the middle of the 20th century, they were usually considered a Finnic tribe akin to the Livonians. This was traditionally the consensus of Estonian historians, while Latvian historians regarded the Curonians as Baltic. Since the 20th century, most sources refer to the Curonians as a Baltic tribe. Some researchers place the Curonians in the eastern Baltic group. However, others hold that the Curonians were related to Old Prussians who belonged in the western Baltic"
  },
  {
    "doc_id": "2-5",
    "content": "Curonians\nwas finally subdued in 1266, when the whole of Courland was partitioned between the Livonian Order and the Archbishop of Riga. The Curonian nobles, among them 40 clans of the descendants of the Curonian kings, who lived in the town of Kuldīga, preserved personal freedom and some of their privileges. Southern Curonians from Megowa, Pilsaten and Ceclis lands gradually assimilated into the nascent Lithuanian nation and ceased to be known as a distinct ethnos by the 16th century. An intense period of Samogitian-Curonian bilingualism is posited because a Curonian linguistic substratum is evident in the Northern Samogitian dialect, an important"
  }
]</tool_response><|im_end|>
<|im_start|>assistant
<think>We need to answer: "What happened to the Curonians who lived in the area where Erich Zakowski was born?" So we need to state that they merged with other Baltic tribes, their language extinct, eventually became part of Lithuanian/ Latvian ethnogenesis; they were conquered by the Livonian Order in 1266, eventually assimilated. Provide citation from doc 2-1 that says they merged with other Baltic tribes, their name given to Courland. Also need to link Zakowski birthplace East Prussia was Curonian area. Need source that East Prussia included Curonian territory. Might need to search "East Prussia Curonians".</think>

<tool_call>
{'name': 'wikipedia_search', 'arguments': {'query': 'East Prussia Curonians', 'top_n': 5}}
</tool_call><|im_end|>
<|im_start|>user
<tool_response>[
  {
    "doc_id": "3-1",
    "content": "Curonians\nthe Kursenieki of the Curonian Spit and the so-called Curonian Kings of Courland. The ethnic origin of the Curonians has been disputed in the past. Until the middle of the 20th century, they were usually considered a Finnic tribe akin to the Livonians. This was traditionally the consensus of Estonian historians, while Latvian historians regarded the Curonians as Baltic. Since the 20th century, most sources refer to the Curonians as a Baltic tribe. Some researchers place the Curonians in the eastern Baltic group. However, others hold that the Curonians were related to Old Prussians who belonged in the western Baltic"
  },
  {
    "doc_id": "3-2",
    "content": "Kursenieki\nreturn to Kursenieki of property confiscated after World War II. Curonians are one of the Baltic tribes. Their culture, religion and architecture are similar to those found in Germany and Sweden. Curonians are related with Lithuanians and Latvians. The Kursenieki were predominantly Lutheran, like most former inhabitants of East Prussia, although some ancient pagan customs were preserved. Most Kursenieki were bilingual or even trilingual: the Curonian language was used within the family and while fishing, German was used in everyday communication, and the language of church services was German and Lithuanian. The Kursenieki were primarily fishermen. Some elements of cuisine"
  },
  {
    "doc_id": "3-3",
    "content": "\"East Prussia\"\nand Prussian Lithuanians decreased over time due to the process of Germanization. The Polish-speaking population concentrated in the south of the province (Masuria and Warmia) and all German geographic atlases at the start of 20th century showed the southern part of East Prussia as Polish with the number of Poles estimated at the time to be 300,000. Kursenieki inhabited the areas around the Curonian lagoon, while Lithuanian-speaking Prussians concentrated in the northeast in (Lithuania Minor). The Old Prussian ethnic group became completely Germanized over time and the Old Prussian language died out in the 18th century. At the beginning of"
  },
  {
    "doc_id": "3-4",
    "content": "\"East Prussia\"\nWhile many fled from the Red Army during the evacuation of East Prussia, Curonians that remained behind were subsequently expelled by the Soviet Union. Only 219 lived along the Curonian Spit in 1955. Many had German names such as Fritz or Hans, a cause for anti-German discrimination. The Soviet authorities considered the Curonians fascists. Because of this discrimination, many immigrated to West Germany in 1958, where the majority of Curonians now live. After the expulsion of the German population ethnic Russians, Belarusians, and Ukrainians were settled in the northern part. In the Soviet part of the region, a policy of"
  },
  {
    "doc_id": "3-5",
    "content": "Kursenieki\nKursenieki The Kursenieki (, – 'Curonians'; ; – 'Prussian Curonians') are a nearly extinct Baltic ethnic group living along the Curonian Spit. \"\"Kuršininkai\"\" refers only to inhabitants of Lithuania and former East Prussia that speak a dialect of Latvian. Some autochthonous inhabitants of Šventoji in Lithuania call themselves \"\"kuršininkai\"\" as well. Kursenieki are often confused with the extinct Curonian Baltic tribe, as neighbouring ethnic groups called Kuršininkai/Kursenieki as \"\"Curonians\"\": in German, Latvian and Lithuanian, Kursenieki and the Curonian tribes are known by the same terms (\"\"Kuren\"\", \"\"kurši\"\" and \"\"kuršiai\"\" respectively). In scientific Lithuanian literature, the name \"\"kuršininkai\"\" is used to"
  }
]</tool_response><|im_end|>
<|im_start|>assistant
<think>Doc 3-4 seems to describe what happened to the Curonians remaining in East Prussia: they were expelled by Soviet Union, considered fascists, many immigrated to West Germany. That addresses the question: "What happened to the Curonians who lived in the area where Erich Zakowski was born?" So answer: They were expelled by the Soviet Union after WWII, considered fascists, and many immigrated to West Germany. Provide citations: doc 1-1 for Zakowski birthplace East Prussia, doc 3-4 for Curonians' fate, doc 2-1 for Curonians background (optional). Need to also show that Curonians lived in that area historically. Doc 3-3 says "Kursenieki inhabited the areas around the Curonian lagoon, while Lithuanian-speaking Prussians concentrated in the northeast in (Lithuania Minor)." That ties Curonians to East Prussia. Use doc 3-3.

Thus answer: They were expelled by the Soviet Union and considered fascists, leading to their diaspora, with many moving to West Germany. Provide rationale.

Let's capture citations:

- Doc 1-1: Erich Zakowski born in East Prussia.
- Doc 3-3: Kursenieki (Curonians) inhabited areas around Curonian lagoon in East Prussia.
- Doc 3-4: Curonian (Kursenieki) people expelled by Soviet Union, considered fascists, many immigrated to West Germany after 1955.

Now produce final answer.</think>

<answer>
```json
{
  "answer": "They were expelled by the Soviet Union and treated as fascists, leading most to immigrate to West Germany.",
  "rationale": "Erich Zakowski was born in East Prussia, the historic region where the Curonian (Kursenieki) tribe lived along the Curonian lagoon. After World War II the remaining Curonians were expelled by the Soviets and many later moved to West Germany.",
  "citations": ["1-1", "3-3", "3-4"]
}
```
</answer><|im_end|>
\end{caseverbatim}
\end{casebox}

\subsection{Well-Justified Abstention}
\label{appendix:abstention_case}

The following case demonstrates how the \texttt{AWA-RL} trained model performs multiple rounds of searching and determines that a question is unanswerable. It then provides a principled abstention, supported by a clear explanation and the necessary citations. The question is also sampled from the MuSiQue validation set.

\begin{casebox}[label={box:abstention_case_1}]{\textbf{How many miles of highways are in the city where the Yongle emperor greeted the person to whom the edict was addressed?} \\ \textbf{Reference Answer:} 140 mi}
\begin{caseverbatim}
<think>We need to parse the question.

"How many miles of highways are in the city where the Yongle emperor greeted the person to whom the edict was addressed?"

We need to identify:

- The Yongle emperor greeted the person to whom the edict was addressed. So we need to know which city that event took place in. Then find how many miles of highways are in that city.

Thus we need to locate: "the Yongle emperor greeted the person to whom the edict was addressed". Likely referencing a historical event: Yongle Emperor (Zhu Di) of Ming dynasty, known for sending expeditions, issuing edicts. "Edict" maybe a "Zheng He" expedition? He greeted person maybe "the King of Java" or "the King of Ceylon"? Edict addressed to someone.

We need to search. Let's search Wikipedia for Yongle emperor greeted person edict.</think>

<tool_call>
{'name': 'wikipedia_search', 'arguments': {'query': 'Yongle emperor greeted the person to whom the edict was addressed', 'top_n': 5}}
</tool_call><|im_end|>
<|im_start|>user
<tool_response>[
  {
    "doc_id": "1-1",
    "content": "\"Yongle Emperor\"\nwho intended to invade China, relations between the Ming Empire and Shakhrukh's state in Persia and Transoxania state considerably improved, and the states exchanged large official delegations on a number of occasions. Both the Ming Empire's envoy to Samarkand and Herat, Chen Cheng, and his counterpart, Ghiyasu'd-Din Naqqah, recorded detailed accounts of their visits to each other's states. One of the Yongle Emperor's consorts was a Jurchen princess, which resulted in many of the eunuchs serving him being of Jurchen origin, notably Yishiha. The Yongle Emperor instituted a Ming governor on Luzon during Zheng He's voyages and appointed Ko-ch'a-lao;"
  },
  {
    "doc_id": "1-2",
    "content": "\"Zhou Youde\"\nZhou Youde Zhou Youde (also romanised as Chou Yu-te; ; died 1680), courtesy name Yichu (), was a Chinese official active in the early Qing dynasty as governor of various provinces. As Viceroy of Liangguang, he contributed significantly towards repealing the Great Evacuation edict, together with Wang Lairen. Widely venerated as a respectable civil servant, or a \"\"benevolent official\"\", Zhou was deified after his death in 1680 and shrines bearing his image were established throughout China. Zhou Youde was born in the late Ming period. He enrolled in the Hongwen Institution in around 1644 and studied prose-editing and poetry,"
  },
  {
    "doc_id": "1-3",
    "content": "\"Xu Wen\"\nTang emissary delivered Emperor Zhuangzong's communique as an \"\"edict\"\" (i.e., treating Wu as a subordinate), Wu officials refused to receive it. Emperor Zhuangzong, not yet wanting to break with Wu, instead resent his communique as a \"\"letter\"\" (i.e., treating Wu as an equal state) that started with the greeting of, \"\"The letter of the emperor of the great Tang to the lord of Wu.\"\" Wu's return communique started with the greeting of, \"\"The respectful letter of the lord of the great Wu to the emperor of the great Tang,\"\" to show respect and yet equality. (Yan's predictions would come true"
  },
  {
    "doc_id": "1-4",
    "content": "Yeongeunmun\nroofed with blue tiles and had \"\"Yeongjomun\"\" (literally gate of welcoming imperial decrees) written on the front tables. In 1539 it was renamed to \"\"Yeongeunmun\"\" because the name of \"\"Yeongjomun\"\" was accused by the Ming dynasty ambassador Xue Tingchong. He claimed that since the imperial envoys carried imperial orders , imperial decrees, and imperial gifts, it was inappropriate to mention only imperial decree. The tablet was written by Zhu Zhifan, who came to Joseon as an ambassador in 1606. As a result of the Treaty of Shimonoseki (1895) which ended the First Sino-Japanese War, Qing China recognized"
  },
  {
    "doc_id": "1-5",
    "content": "\"Sino-Tibetan relations during the Ming dynasty\"\n1398–1402), the Yongle Emperor was aided by the Buddhist monk Yao Guangxiao, and like his father, the Hongwu Emperor, the Yongle Emperor was \"\"well-disposed towards Buddhism\"\", claims Rossabi. On March 10, 1403, the Yongle Emperor invited Deshin Shekpa, 5th Karmapa Lama (1384–1415), to his court, even though the fourth Karmapa had rejected the invitation of the Hongwu Emperor. A Tibetan translation in the 16th century preserves the letter of the Yongle Emperor, which the Association for Asian Studies notes is polite and complimentary towards the Karmapa. The letter of invitation reads, In order to seek out the Karmapa, the Yongle"
  }
]</tool_response><|im_end|>
<|im_start|>assistant
<think>Result 1-5 mentions Yongle Emperor invited Deshin Shekpa, 5th Karmapa Lama. Edict? The person to whom the edict was addressed? The edict maybe the invitation letter to the Karmapa. So the city where the Yongle emperor greeted the person (the Karmapa Lama) - where did he greet? He invited him to his court, maybe at the capital, Beijing. But they might have met at the "court" in Beijing. Need to verify.

Open doc 1-5.</think>

<tool_call>
{'name': 'wikipedia_search', 'arguments': {'query': 'Yongle Emperor greeted Deshin Shekpa city meeting', 'top_n': 5}}
</tool_call><|im_end|>
<|im_start|>user
<tool_response>[
  {
    "doc_id": "2-1",
    "content": "\"Yongle Emperor\"\nhad many of the best scholars chosen as candidates and took great care in choosing them, even creating terms by which he hired people. He was also concerned about the degeneration of Buddhism in China. Tibetan Buddhism was patronised by Yongle. In 1403, the Yongle Emperor sent messages, gifts, and envoys to Tibet inviting Deshin Shekpa, the fifth Gyalwa Karmapa of the Kagyu school of Tibetan Buddhism, to visit the imperial capital – apparently after having a vision of the bodhisattva Avalokitesvara. After a long journey, Deshin Shekpa arrived in Nanjing on 10 April 1407 riding on an elephant towards"
  },
  {
    "doc_id": "2-2",
    "content": "\"Deshin Shekpa, 5th Karmapa Lama\"\nbecause Emperor Zhu Di, the Yongle Emperor, (1402–1424) had a vision of Avalokitesvara. He also required religious ceremonies to be held for his deceased parents. After a long journey beginning in 1403, he arrived in Nanjing, the then capital on April 10, 1407 on an elephant, at the imperial palace, where tens of thousands of monks greeted him. He convinced the emperor that there were different Buddhist branches for different people and that does not mean that one branch is better than the other. The Karmapa was very well received during his visit to the capital and a number of"
  },
  {
    "doc_id": "2-3",
    "content": "\"Yongle Emperor\"\nthe imperial palace, where tens of thousands of monks greeted him. Deshin Shekpa convinced the Yongle Emperor that there were different religions for different people, which does not mean that one is better than the others. The Karmapa was very well received during his visit and a number of miraculous occurrences were reported. He also performed ceremonies for the imperial family. The emperor presented him with 700 measures of silver objects and bestowed the title of 'Precious Religious King, Great Loving One of the West, Mighty Buddha of Peace'. Aside from the religious matter, the Yongle Emperor wished to establish"
  },
  {
    "doc_id": "2-4",
    "content": "\"Yongle Emperor\"\nan alliance with the Karmapa similar to the one the 13th- and 14th-century Yuan khans had established with the Sakyapa. He apparently offered to send armies to unify Tibet under the Karmapa but Deshin Shekpa demurred, as parts of Tibet were still firmly controlled by partisans of the former Yuan dynasty. Deshin Shekpa left Nanjing on 17 May 1408. In 1410, he returned to Tsurphu where he had his monastery rebuilt following severe damage from an earthquake. When it was time for him to choose an heir, the Yongle Emperor wanted to choose his second son, Zhu Gaoxu. Zhu Gaoxu"
  },
  {
    "doc_id": "2-5",
    "content": "\"Sino-Tibetan relations during the Ming dynasty\"\n1398–1402), the Yongle Emperor was aided by the Buddhist monk Yao Guangxiao, and like his father, the Hongwu Emperor, the Yongle Emperor was \"\"well-disposed towards Buddhism\"\", claims Rossabi. On March 10, 1403, the Yongle Emperor invited Deshin Shekpa, 5th Karmapa Lama (1384–1415), to his court, even though the fourth Karmapa had rejected the invitation of the Hongwu Emperor. A Tibetan translation in the 16th century preserves the letter of the Yongle Emperor, which the Association for Asian Studies notes is polite and complimentary towards the Karmapa. The letter of invitation reads, In order to seek out the Karmapa, the Yongle"
  }
]</tool_response><|im_end|>
<|im_start|>assistant
<think>Thus the city is Nanjing (then imperial capital). The question: "How many miles of highways are in the city where the Yongle emperor greeted the person to whom the edict was addressed?" So we need to find "highways miles" statistic for Nanjing.

Search for "Nanjing highways miles".</think>

<tool_call>
{'name': 'wikipedia_search', 'arguments': {'query': 'Nanjing highways mileage', 'top_n': 5}}
</tool_call><|im_end|>
<|im_start|>user
<tool_response>[
  {
    "doc_id": "3-1",
    "content": "Nanjing\nopened for public service in 2011. As an important regional hub in the Yangtze River Delta, Nanjing is well-connected by over 60 state and provincial highways to all parts of China. Express highways such as Hu–Ning, Ning–He, Ning–Hang enable commuters to travel to Shanghai, Hefei, Hangzhou, and other important cities quickly and conveniently. Inside the city of Nanjing, there are of highways, with a highway coverage density of 3.38 kilometers per hundred square kilometers (5.44 mi/100 sq mi). The total road coverage density of the city is 112.56 kilometers per hundred square kilometers (181.15 mi/100 sq mi). The two artery"
  },
  {
    "doc_id": "3-2",
    "content": "\"Shanghai–Nanjing Expressway\"\nShanghai–Nanjing Expressway The Shanghai–Nanjing Expressway () is a modern, enclosed, four-lane, two-way highway. Each lane is wide, and the highway's roadbed is wide. There is a dividing strip in the center of the highway and a emergency parking area on each side of the highway. The speed limit is . It is in length. It runs from Zhenru, Shanghai to Maqun, Nanjing via Anting, Kunshan, Suzhou, Shuofang (), Wuxi, Changzhou, Danyang, Zhenjiang and Jurong. The Shanghai section is in length, and the Jiangsu section is long. The Zhenjiang section (a branch highway) is long. In Shanghai, the highway is coterminous"
  },
  {
    "doc_id": "3-3",
    "content": "Nanjing\nroads in Nanjing are Zhongshan Road and Hanzhong. The two roads cross in the city center, Xinjiekou. Expressways: National Highway (GXXX): The city also boasts an efficient network of public transportation, which mainly consists of bus, taxi and metro systems. The bus network, which is currently run by three companies since 2011, provides more than 370 routes covering all parts of the city and suburban areas. At present, the Nanjing Metro system has a grand total of of route and 173 stations across 10 lines. They are Line 1, Line 2, Line 3, Line 4, Line 10, Line S1, Line"
  },
  {
    "doc_id": "3-4",
    "content": "\"Nanjing Road\"\nfranchised stores were also set up at that time. On 23 August 1937 a bomb was dropped on Nanjing Road by a Chinese plane in an attempt to lighten its load while being pursued by Japanese planes. The bomb hit two department stores, killing 612 people and injuring another 482. In 2000, as a part of the development plan held by the local government, Nanjing Road was renovated to be a characteristic pedestrian street. The width is about 28 meters and the total length is 1,200 meters, which extends from Middle He'nan Road to Middle Xizang Road. In 2007, the"
  },
  {
    "doc_id": "3-5",
    "content": "Nanjing\nS3, Line S7, Line S8 and Line S9. The city is planning to complete a 17-line Metro and light-rail system by 2030. The expansion of the Metro network will greatly facilitate the intracity transportation and reduce the currently heavy traffic congestion. Nanjing's airport, Lukou International Airport, serves both national and international flights. In 2013, Nanjing airport handled 15,011,792 passengers and 255,788.6 tonnes of freight. The airport currently has 85 routes to national and international destinations, which include Japan, Korea, Thailand, Malaysia, Singapore, United States and Germany. The airport is connected by a 29-kilometer (18 mi) highway directly to the city"
  }
]</tool_response><|im_end|>
<|im_start|>assistant
<think>Doc 3-1 says: "Inside the city of Nanjing, there are of highways". The content cut off; we need the exact number. Let's open more lines maybe? The snippet truncated. Could be missing due to formatting. Might need to rely on known data: I recall that Nanjing has about 600 km of highways? Actually 60 state and provincial highways. The doc says "over 60 state and provincial highways". Not miles though.

But the question asks "How many miles of highways are in the city". Might refer to "highways miles" as in "length of highways within the city" maybe in the Wikipedia article "Nanjing" includes "total length of expressways" maybe 250 km? Let's open the Nanjing Wikipedia page directly.</think>

<tool_call>
{'name': 'wikipedia_search', 'arguments': {'query': 'Nanjing city highways total length km', 'top_n': 5}}
</tool_call><|im_end|>
<|im_start|>user
<tool_response>[
  {
    "doc_id": "4-1",
    "content": "Nanjing\nopened for public service in 2011. As an important regional hub in the Yangtze River Delta, Nanjing is well-connected by over 60 state and provincial highways to all parts of China. Express highways such as Hu–Ning, Ning–He, Ning–Hang enable commuters to travel to Shanghai, Hefei, Hangzhou, and other important cities quickly and conveniently. Inside the city of Nanjing, there are of highways, with a highway coverage density of 3.38 kilometers per hundred square kilometers (5.44 mi/100 sq mi). The total road coverage density of the city is 112.56 kilometers per hundred square kilometers (181.15 mi/100 sq mi). The two artery"
  },
  {
    "doc_id": "4-2",
    "content": "\"Shanghai–Nanjing Expressway\"\nShanghai–Nanjing Expressway The Shanghai–Nanjing Expressway () is a modern, enclosed, four-lane, two-way highway. Each lane is wide, and the highway's roadbed is wide. There is a dividing strip in the center of the highway and a emergency parking area on each side of the highway. The speed limit is . It is in length. It runs from Zhenru, Shanghai to Maqun, Nanjing via Anting, Kunshan, Suzhou, Shuofang (), Wuxi, Changzhou, Danyang, Zhenjiang and Jurong. The Shanghai section is in length, and the Jiangsu section is long. The Zhenjiang section (a branch highway) is long. In Shanghai, the highway is coterminous"
  },
  {
    "doc_id": "4-3",
    "content": "Nanjing\nroads in Nanjing are Zhongshan Road and Hanzhong. The two roads cross in the city center, Xinjiekou. Expressways: National Highway (GXXX): The city also boasts an efficient network of public transportation, which mainly consists of bus, taxi and metro systems. The bus network, which is currently run by three companies since 2011, provides more than 370 routes covering all parts of the city and suburban areas. At present, the Nanjing Metro system has a grand total of of route and 173 stations across 10 lines. They are Line 1, Line 2, Line 3, Line 4, Line 10, Line S1, Line"
  },
  {
    "doc_id": "4-4",
    "content": "\"Nanjing Road\"\nfranchised stores were also set up at that time. On 23 August 1937 a bomb was dropped on Nanjing Road by a Chinese plane in an attempt to lighten its load while being pursued by Japanese planes. The bomb hit two department stores, killing 612 people and injuring another 482. In 2000, as a part of the development plan held by the local government, Nanjing Road was renovated to be a characteristic pedestrian street. The width is about 28 meters and the total length is 1,200 meters, which extends from Middle He'nan Road to Middle Xizang Road. In 2007, the"
  },
  {
    "doc_id": "4-5",
    "content": "Nanjing\nS3, Line S7, Line S8 and Line S9. The city is planning to complete a 17-line Metro and light-rail system by 2030. The expansion of the Metro network will greatly facilitate the intracity transportation and reduce the currently heavy traffic congestion. Nanjing's airport, Lukou International Airport, serves both national and international flights. In 2013, Nanjing airport handled 15,011,792 passengers and 255,788.6 tonnes of freight. The airport currently has 85 routes to national and international destinations, which include Japan, Korea, Thailand, Malaysia, Singapore, United States and Germany. The airport is connected by a 29-kilometer (18 mi) highway directly to the city"
  }
]</tool_response><|im_end|>
<|im_start|>assistant
<think>The doc still missing the number. Possibly the article uses placeholder. Could be "Inside the city of Nanjing, there are 338 km of highways". Not sure.

But the question asks "How many miles of highways are in the city". We could answer with the number of miles (converted from km). If we have km, we can compute miles (1 km = 0.621371 mi). So maybe answer is about 380 miles.

Better to find a reliable source: maybe "Nanjing has 338 km (210 mi) of highways". Let's search web but limited to Wikipedia. Could be in "List of cities by highway density". However we can rely on doc 3-1 which says "highway coverage density of 3.38 kilometers per hundred square kilometers (5.44 mi/100 sq mi)". Not total length.

Alternative: maybe the city is not Nanjing but Beijing (since Yongle Emperor moved capital to Beijing). He greeted the person (maybe the envoys) at the "Imperial Court in Beijing". Edict to "all subjects". But the city's highways mileage may be easier to find: e.g., "Beijing has 800 miles of highways". Not sure.

Let's verify Yongle Emperor greeting person to whom the edict was addressed. Maybe referencing "the Yongle Emperor issued an edict to the Kingdom of Malacca; he greeted the envoy there." Another angle: The Yongle Emperor sent a "edict" (imperial decree) to "the King of Annam" (Vietnam). He greeted the person (the Vietnamese envoy) at the capital Beijing. The city would be Beijing. Need to confirm.

Search "Yongle emperor edict to Kingdom of Malacca".</think>

<tool_call>
{'name': 'wikipedia_search', 'arguments': {'query': 'Yongle Emperor edict to the Kingdom of Malacca', 'top_n': 5}}
</tool_call><|im_end|>
<|im_start|>user
<tool_response>[
  {
    "doc_id": "5-1",
    "content": "\"Parameswara (king)\"\ncentre of trade and commerce began to spread all over the eastern part of the world and reached as far as China. The Yongle Emperor of the Ming dynasty, who reigned from 1402 till 1424, sent his envoy known as Yin Qing to Malacca in 1405. Yin Qing's visit opened the way for the establishment of friendly relations between Malacca and China. Chinese merchants began calling at the port of Malacca, joining other foreign traders notably the Javanese, Indians, Chinese, and Burmese who came to establish their trading bases and settle in Malacca, soaring its population to 2000 during Parameswara's"
  },
  {
    "doc_id": "5-2",
    "content": "\"Parameswara (king)\"\nreign. In 1411, Parameswara, his wife, his son, and a royal party of 540 people left for China with Admiral Zheng He to pay homage to the Yongle Emperor. Yongle praised Parameswara and acknowledged him as the rightful ruler of Malacca. He then presented Parameswara with a seal, silk and a yellow umbrella as symbols of royalty and also a letter appointing Parameswara as the ruler of Malacca. Malacca was then recognised as a kingdom by the Emperor of China. The envoy returned to Malacca together with a fleet led by Zheng He. The Chinese chronicles mention that in 1414,"
  },
  {
    "doc_id": "5-3",
    "content": "\"Malacca Sultanate\"\ntreasury and supply were stored. The growth of Malacca coincided with the rising power of Ayuthaya in the north. The growing ambitions of the kingdom against its neighbours and Malay Peninsula had alarmed the ruler of Malacca. In a preemptive measure, the king headed a royal visit to China in 1418 to raise his concerns about the threat. Yongle responded in October 1419 by sending his envoy to warn the Siamese ruler. Relationship between the China and Malacca were further strengthened by several envoys to China, led by the Malaccan princes in the years 1420, 1421 and 1423. Due to"
  },
  {
    "doc_id": "5-4",
    "content": "\"Malacca Sultanate\"\nMalacca, Parameswara initiated the development of the place and laid the foundation of a trade port. The indigenous inhabitants of the straits, the Orang Laut, were employed to patrol the adjacent sea areas, to repel other petty pirates, and to direct traders to Malacca. Within years, news about Malacca becoming a centre of trade and commerce began to spread all over the eastern part of the world. In 1405, Yongle Emperor of Ming Dynasty (r. 1402–1424) sent his envoy headed by Yin Qing to Malacca. Yin Qing's visit opened the way for the establishment of friendly relations between Malacca and"
  },
  {
    "doc_id": "5-5",
    "content": "\"Yongle Emperor\"\nXu Chailao) to that position in 1405. China also had vassals among the leaders in the archipelago. China attained ascendancy in trade with the area in the Yongle Emperor's reign. The local rulers on Luzon were \"\"confirmed\"\" by the governor or \"\"high officer\"\" appointed by the Yongle Emperor. States in Luzon, Sulu (under King Paduka Pahala), Sumatra, and Brunei all established diplomatic relations with the Ming Empire and exchanged envoys and sent tribute to the Yongle Emperor. The Yongle Emperor exchanged ambassadors with Shahrukh Mirza, sending Chen Cheng to Samarkand and Herat, and Shahrukh sent Ghiyāth al-dīn Naqqāsh to Beijing."
  }
]</tool_response><|im_end|>
<|im_start|>assistant
<think>But the question: "the city where the Yongle emperor greeted the person to whom the edict was addressed". Edict was addressed to Parameswara (King of Malacca) maybe? The emperor greeted the person (Parameswara) at his court in Beijing? Actually Parameswara traveled to China to pay homage; he met Yongle Emperor in Beijing. So city is Beijing.

Now we need "How many miles of highways are in the city" for Beijing. Let's search "Beijing highways miles".</think>

<tool_call>
{'name': 'wikipedia_search', 'arguments': {'query': 'Beijing highways mileage', 'top_n': 5}}
</tool_call><|im_end|>
<|im_start|>user
<tool_response>[
  {
    "doc_id": "6-1",
    "content": "Beijing\ntraffic. The Fengtai Railway Station has been closed for renovation. In outlying suburbs and counties of Beijing, there are over 40 railway stations. From Beijing, direct passenger train service is available to most large cities in China. International train service is available to Mongolia, Russia, Vietnam and North Korea. Passenger trains in China are numbered according to their direction in relation to Beijing. Beijing is connected by road links to all parts of China as part of the National Trunk Road Network. Nine expressways of China serve Beijing, as do eleven China National Highways. Beijing's urban transport is dependent upon"
  },
  {
    "doc_id": "6-2",
    "content": "\"Expressways of Beijing\"\nbeen abandoned as of Yanjiao, Hebei. It is a short and rather old expressway. Route: Beiguan Roundabout - E. 6th Ring Road - Baimiao - Yanjiao - China National Highway 102 Speed Limit: throughout Tolls: As of E. 6th Ring Road, heading east The Jingshen Expressway to Shenyang runs for 658 kilometres and is one of the longest expressways to leave Beijing. Route: Sifang Bridge (E. 4th Ring Road) - Wufang Bridge (E. 5th Ring Road) - Bailu Toll Gate - E. 6th Ring Road - Xianghe (Hebei) - Baodi (Tianjin) - Yutian - Lulong - Beidaihe Area - Shanhaiguan"
  },
  {
    "doc_id": "6-3",
    "content": "\"Beijing Bus\"\nBeijing Bus Public bus service in Beijing is the among the most extensive, widely used and affordable form of public transportation in urban and suburban districts of the city. In 2015, the entire network consisted of 876 routes with a fleet of 24,347 buses and trolleybuses carried 3.98 billion passengers annually. Trolleybuses run on over 29 routes including 6, 38, 42, 65, 101-112, 114-118, 124, 128, 301, BRT 1-3. Many of these trolleybus routes are located inside the Third Ring Road but some, such as 301 and BRT 1-3, do extend as far out as the Fifth Ring Road. Since"
  },
  {
    "doc_id": "6-4",
    "content": "\"Beijing Subway\"\nBeijing Subway The Beijing Subway is a rapid transit rail network that serves the urban and suburban districts of Beijing municipality. The subway is the world's busiest in annual ridership, with 3.78 billion trips delivered in 2017, averaging 10.35 million per day, with peak single-day ridership reaching 13.49 million. The subway network has 22 lines, 370 stations and (if not counting Xijiao line, the result is ) of route length in operation, and is the second longest subway system in the world after the Shanghai Metro. The Beijing Subway opened in 1969 and is the oldest metro system in mainland"
  },
  {
    "doc_id": "6-5",
    "content": "\"China National Highways\"\n\"\"National Highway Network Planning (2013 - 2030)\"\", which will bring the total number of highways to 119, with 81 connecting highways between them. The total mileage will be increased to , with increased focus on the western and less developed regions. Although an accepted speed limit on China National Highways is 100 km/h, it is common to see cars pass other by in well in excess of 100 km/h. The lack of speed detection cameras on some routes, lack of a posted speed limit outside of localities and tremendous distances of road mileage, means that enforcing this speed limit is"
  }
]</tool_response><|im_end|>
<|im_start|>assistant
<think>The doc does not give explicit miles.

Maybe easier: The city could be "Nanjing" as the imperial capital where Yongle greeted the Karmapa (person maybe not edict). Edict was the invitation (the emperor sent message). So the person addressed to whom the edict was addressed would be Deshin Shekpa, the Karmapa Lama. The Yongle emperor greeted Deshin Shekpa in Nanjing (the city). So answer: "Inside the city of Nanjing, there are 338 km of highways" which is about 210 miles. Provide citation doc 3-1 (or 4-1) which says "3.38 kilometers per hundred square kilometers (5.44 mi/100 sq mi)" but not total length. However we could compute that total highways maybe 338 km = 210 miles. Might be approximate.

But we need explicit miles number. Could be "over 60 state and provincial highways" which is not a length but count. Not miles.

Thus maybe answer: "not enough information". Because we cannot get a precise miles figure.

Thus we should answer "not enough information". Provide rationale: we identified city (Nanjing) but the Wikipedia excerpt does not give a specific length in miles for highways.

Thus final answer: not enough information.

Let's verify that we have citation for city identification: doc 2-1 shows Yongle Emperor invited Deshin Shekpa to imperial capital, which was Nanjing. Also doc 3-1 shows highways info about Nanjing.

Thus we can answer not enough info.

</think>

<answer>
```json
{
  "answer": "not enough information",
  "rationale": "The Yongle Emperor’s greeting to the person addressed in the edict (the 5th Karmapa Lama Deshin Shekpa) took place in Nanjing, as the emperor invited him to the imperial capital. However, the Wikipedia excerpt for Nanjing provides the highway coverage density (3.38 km/100 km²) and the total number of state and provincial highways (over 60), but it does not specify a total length in miles for the highways within the city. Without a precise mileage figure, the question cannot be answered definitively.",
  "citations": ["2-1", "3-1"]
}
```
</answer><|im_end|>
\end{caseverbatim}
\end{casebox}

\newpage

\section{Full Experimental Results}
\label{appendix:full_results}
The complete aggregated results across all configurations are summarized in Table \ref{tab:full_results}. Additionally, the detailed results for MuSiQue, 2WikiMultihopQA, and HotpotQA can be found in Table \ref{tab:musique_results}, Table \ref{tab:2wiki_results}, and Table \ref{tab:hotpotqa_results}, respectively.

\begin{table*}[h]
\centering
\resizebox{\textwidth}{!}{%
\begin{tabular}{ll ccc cccc}
\toprule
\textbf{Paradigm} & \textbf{Setting / Initialization} & \textbf{Ratio (\%)} & \boldmath{$r_{ref}$} & \boldmath{$\gamma$} & \textbf{Acc} & \textbf{Refuse} & \textbf{Prec} & \textbf{\texttt{RA-F1}} \\
\midrule

\multirow{12}{*}{\shortstack[l]{\textbf{Data-driven}\\\textbf{Baselines}}} 
& SFT with refusal trajectories & 0 & - & - & 0.530 & 0.000 & 0.530 & 0.530 \\
& & 10 & - & - & 0.525 & 0.109 & 0.593 & 0.555 \\
& & 20 & - & - & 0.512 & 0.131 & 0.595 & 0.548 \\
& & 30 & - & - & 0.505 & 0.107 & 0.570 & 0.534 \\
\cmidrule{2-9}
& RL from scratch (Unanswerable) & 0 & - & - & 0.472 & 0.005 & 0.474 & 0.473 \\
& & 10 & - & - & 0.483 & 0.060 & 0.510 & 0.496 \\
& & 20 & - & - & 0.425 & 0.293 & 0.578 & 0.487 \\
& & 30 & - & - & 0.305 & 0.522 & 0.588 & 0.394 \\
\cmidrule{2-9}
& Cold-start RL (Unanswerable) & 0 & - & - & 0.544 & 0.002 & 0.545 & 0.545 \\
& & 10 & - & - & 0.545 & 0.001 & 0.545 & 0.545 \\
& & 20 & - & - & 0.551 & 0.004 & 0.553 & 0.552 \\
& & 30 & - & - & 0.557 & 0.018 & 0.567 & 0.562 \\
\midrule

\multirow{9}{*}{\shortstack[l]{\textbf{Static Reward}\\\textbf{Baselines}}}
& RL from scratch ($r_{ref}$) & - & 0.00 & - & 0.493 & 0.017 & 0.500 & 0.496 \\
& & - & 0.05 & - & 0.000 & 0.999 & 0.000 & 0.000 \\
& & - & 0.10 & - & 0.000 & 0.999 & 0.000 & 0.000 \\
& & - & 0.20 & - & 0.000 & 0.921 & 0.000 & 0.000 \\
\cmidrule{2-9}
& Cold-start RL ($r_{ref}$) & - & 0.00 & - & 0.552 & 0.012 & 0.559 & 0.556 \\
& & - & 0.20 & - & 0.488 & 0.239 & 0.649 & 0.553 \\
& & - & 0.40 & - & 0.469 & 0.338 & 0.708 & 0.561 \\
& & - & 0.60 & - & 0.438 & 0.421 & 0.750 & 0.550 \\
& & - & 0.80 & - & 0.426 & 0.441 & 0.747 & 0.540 \\
\midrule

\multirow{12}{*}{\shortstack[l]{\textbf{\texttt{AWA-RL}}\\\textbf{(Ours)}}}
& RL from scratch ($\gamma$) & - & - & 1.000 & 0.322 & 0.360 & 0.475 & 0.381 \\
& & - & - & 0.500 & 0.381 & 0.395 & 0.605 & 0.459 \\
& & - & - & 0.250 & 0.441 & 0.269 & 0.581 & 0.499 \\
& & - & - & 0.200 & 0.459 & 0.246 & 0.591 & \textbf{0.515} \\
& & - & - & 0.125 & 0.487 & 0.104 & 0.537 & 0.511 \\
& & - & - & 0.000 & 0.493 & 0.017 & 0.500 & 0.496 \\
\cmidrule{2-9}
& Cold-start RL ($\gamma$) & - & - & 1.000 & 0.458 & 0.383 & 0.738 & 0.561 \\
& & - & - & 0.500 & 0.484 & 0.306 & 0.699 & 0.568 \\
& & - & - & 0.250 & 0.509 & 0.225 & 0.664 & 0.573 \\
& & - & - & 0.200 & 0.529 & 0.192 & 0.662 & \textbf{0.585} \\
& & - & - & 0.125 & 0.545 & 0.125 & 0.628 & 0.582 \\
& & - & - & 0.000 & 0.552 & 0.012 & 0.559 & 0.556 \\
\bottomrule
\end{tabular}%
}
\caption{Comprehensive experimental results across all training paradigms, settings, and hyperparameters. We report Accuracy (Acc), Refusal Rate (Refuse), Precision (Prec), and Reliability-Aware F1 (\texttt{RA-F1}). The best performing configurations for our proposed \texttt{AWA-RL} in both from-scratch and cold-start settings are highlighted in bold. Dashes (-) indicate that a parameter is not applicable to the specific setting.}
\label{tab:full_results}
\end{table*}

\begin{table*}[h]
\centering
\resizebox{\textwidth}{!}{%
\begin{tabular}{ll ccc cccc}
\toprule
\textbf{Paradigm} & \textbf{Setting / Initialization} & \textbf{Ratio (\%)} & \boldmath{$r_{ref}$} & \boldmath{$\gamma$} & \textbf{Acc} & \textbf{Refuse} & \textbf{Prec} & \textbf{\texttt{RA-F1}} \\
\midrule

\multirow{12}{*}{\shortstack[l]{\textbf{Data-driven}\\\textbf{Baselines}}} 
& SFT with refusal trajectories & 0 & - & - & 0.377 & 0.000 & 0.377 & 0.377 \\
& & 10 & - & - & 0.371 & 0.058 & 0.394 & 0.382 \\
& & 20 & - & - & 0.367 & 0.066 & 0.393 & 0.380 \\
& & 30 & - & - & 0.362 & 0.050 & 0.381 & 0.371 \\
\cmidrule{2-9}
& RL from scratch (Unanswerable) & 0 & - & - & 0.274 & 0.010 & 0.277 & 0.275 \\
& & 10 & - & - & 0.272 & 0.096 & 0.301 & 0.286 \\
& & 20 & - & - & 0.213 & 0.471 & 0.403 & 0.279 \\
& & 30 & - & - & 0.094 & 0.751 & 0.378 & 0.151 \\
\cmidrule{2-9}
& Cold-start RL (Unanswerable) & 0 & - & - & 0.378 & 0.001 & 0.378 & 0.378 \\
& & 10 & - & - & 0.373 & 0.000 & 0.373 & 0.373 \\
& & 20 & - & - & 0.367 & 0.003 & 0.368 & 0.368 \\
& & 30 & - & - & 0.373 & 0.017 & 0.379 & 0.376 \\
\midrule

\multirow{9}{*}{\shortstack[l]{\textbf{Static Reward}\\\textbf{Baselines}}}
& RL from scratch ($r_{ref}$) & - & 0.00 & - & 0.279 & 0.025 & 0.286 & 0.283 \\
& & - & 0.05 & - & 0.000 & 1.000 & 0.000 & 0.000 \\
& & - & 0.10 & - & 0.000 & 0.999 & 0.000 & 0.000 \\
& & - & 0.20 & - & 0.000 & 0.927 & 0.000 & 0.000 \\
\cmidrule{2-9}
& Cold-start RL ($r_{ref}$) & - & 0.00 & - & 0.392 & 0.003 & 0.393 & 0.393 \\
& & - & 0.20 & - & 0.339 & 0.195 & 0.421 & 0.376 \\
& & - & 0.40 & - & 0.324 & 0.364 & 0.509 & 0.396 \\
& & - & 0.60 & - & 0.304 & 0.483 & 0.588 & 0.401 \\
& & - & 0.80 & - & 0.281 & 0.507 & 0.570 & 0.376 \\
\midrule

\multirow{12}{*}{\shortstack[l]{\textbf{\texttt{AWA-RL}}\\\textbf{(Ours)}}}
& RL from scratch ($\gamma$) & - & - & 1.000 & 0.121 & 0.550 & 0.269 & 0.167 \\
& & - & - & 0.500 & 0.167 & 0.642 & 0.466 & 0.246 \\
& & - & - & 0.250 & 0.212 & 0.437 & 0.377 & 0.271 \\
& & - & - & 0.200 & 0.246 & 0.387 & 0.401 & \textbf{0.305} \\
& & - & - & 0.125 & 0.278 & 0.155 & 0.329 & 0.301 \\
& & - & - & 0.000 & 0.279 & 0.025 & 0.286 & 0.283 \\
\cmidrule{2-9}
& Cold-start RL ($\gamma$) & - & - & 1.000 & 0.321 & 0.433 & 0.566 & 0.410 \\
& & - & - & 0.500 & 0.345 & 0.307 & 0.498 & 0.408 \\
& & - & - & 0.250 & 0.378 & 0.179 & 0.460 & 0.415 \\
& & - & - & 0.200 & 0.395 & 0.137 & 0.458 & \textbf{0.424} \\
& & - & - & 0.125 & 0.395 & 0.076 & 0.427 & 0.411 \\
& & - & - & 0.000 & 0.392 & 0.003 & 0.393 & 0.393 \\
\bottomrule
\end{tabular}%
}
\caption{Detailed experimental results on the \textbf{MuSiQue} dataset across all training paradigms and hyperparameters. We report Accuracy (Acc), Refusal Rate (Refuse), Precision (Prec), and Reliability-Aware F1 (\texttt{RA-F1}). The best performing configurations for \texttt{AWA-RL} in both from-scratch and cold-start settings are highlighted in bold. Dashes (-) indicate not applicable.}
\label{tab:musique_results}
\end{table*}

\begin{table*}[h]
\centering
\resizebox{\textwidth}{!}{%
\begin{tabular}{ll ccc cccc}
\toprule
\textbf{Paradigm} & \textbf{Setting / Initialization} & \textbf{Ratio (\%)} & \boldmath{$r_{ref}$} & \boldmath{$\gamma$} & \textbf{Acc} & \textbf{Refuse} & \textbf{Prec} & \textbf{\texttt{RA-F1}} \\
\midrule

\multirow{12}{*}{\shortstack[l]{\textbf{Data-driven}\\\textbf{Baselines}}} 
& SFT with refusal trajectories & 0 & - & - & 0.515 & 0.000 & 0.515 & 0.515 \\
& & 10 & - & - & 0.512 & 0.228 & 0.663 & 0.578 \\
& & 20 & - & - & 0.490 & 0.268 & 0.669 & 0.566 \\
& & 30 & - & - & 0.489 & 0.232 & 0.637 & 0.553 \\
\cmidrule{2-9}
& RL from scratch (Unanswerable) & 0 & - & - & 0.521 & 0.003 & 0.523 & 0.522 \\
& & 10 & - & - & 0.534 & 0.044 & 0.559 & 0.546 \\
& & 20 & - & - & 0.482 & 0.222 & 0.620 & 0.542 \\
& & 30 & - & - & 0.369 & 0.449 & 0.670 & 0.476 \\
\cmidrule{2-9}
& Cold-start RL (Unanswerable) & 0 & - & - & 0.595 & 0.004 & 0.597 & 0.596 \\
& & 10 & - & - & 0.596 & 0.002 & 0.597 & 0.597 \\
& & 20 & - & - & 0.610 & 0.007 & 0.614 & 0.612 \\
& & 30 & - & - & 0.611 & 0.033 & 0.632 & 0.621 \\
\midrule

\multirow{9}{*}{\shortstack[l]{\textbf{Static Reward}\\\textbf{Baselines}}}
& RL from scratch ($r_{ref}$) & - & 0.00 & - & 0.563 & 0.013 & 0.570 & 0.567 \\
& & - & 0.05 & - & 0.000 & 0.999 & 0.000 & 0.000 \\
& & - & 0.10 & - & 0.000 & 0.999 & 0.000 & 0.000 \\
& & - & 0.20 & - & 0.000 & 0.895 & 0.000 & 0.000 \\
\cmidrule{2-9}
& Cold-start RL ($r_{ref}$) & - & 0.00 & - & 0.591 & 0.030 & 0.609 & 0.600 \\
& & - & 0.20 & - & 0.467 & 0.400 & 0.778 & 0.584 \\
& & - & 0.40 & - & 0.457 & 0.447 & 0.826 & 0.589 \\
& & - & 0.60 & - & 0.425 & 0.497 & 0.845 & 0.566 \\
& & - & 0.80 & - & 0.420 & 0.509 & 0.835 & 0.559 \\
\midrule

\multirow{12}{*}{\shortstack[l]{\textbf{\texttt{AWA-RL}}\\\textbf{(Ours)}}}
& RL from scratch ($\gamma$) & - & - & 1.000 & 0.402 & 0.223 & 0.517 & 0.452 \\
& & - & - & 0.500 & 0.462 & 0.233 & 0.602 & 0.523 \\
& & - & - & 0.250 & 0.533 & 0.155 & 0.631 & 0.578 \\
& & - & - & 0.200 & 0.551 & 0.163 & 0.658 & \textbf{0.600} \\
& & - & - & 0.125 & 0.559 & 0.074 & 0.604 & 0.580 \\
& & - & - & 0.000 & 0.563 & 0.013 & 0.570 & 0.567 \\
\cmidrule{2-9}
& Cold-start RL ($\gamma$) & - & - & 1.000 & 0.439 & 0.476 & 0.838 & 0.576 \\
& & - & - & 0.500 & 0.465 & 0.429 & 0.814 & 0.592 \\
& & - & - & 0.250 & 0.480 & 0.380 & 0.774 & 0.593 \\
& & - & - & 0.200 & 0.508 & 0.341 & 0.771 & 0.612 \\
& & - & - & 0.125 & 0.545 & 0.244 & 0.721 & \textbf{0.621} \\
& & - & - & 0.000 & 0.591 & 0.030 & 0.609 & 0.600 \\
\bottomrule
\end{tabular}%
}
\caption{Detailed experimental results on the \textbf{2Wiki} dataset across all training paradigms and hyperparameters. We report Accuracy (Acc), Refusal Rate (Refuse), Precision (Prec), and Reliability-Aware F1 (\texttt{RA-F1}). The best performing configurations for \texttt{AWA-RL} in both from-scratch and cold-start settings are highlighted in bold. Dashes (-) indicate not applicable.}
\label{tab:2wiki_results}
\end{table*}

\begin{table*}[h]
\centering
\resizebox{\textwidth}{!}{%
\begin{tabular}{ll ccc cccc}
\toprule
\textbf{Paradigm} & \textbf{Setting / Initialization} & \textbf{Ratio (\%)} & \boldmath{$r_{ref}$} & \boldmath{$\gamma$} & \textbf{Acc} & \textbf{Refuse} & \textbf{Prec} & \textbf{\texttt{RA-F1}} \\
\midrule

\multirow{12}{*}{\shortstack[l]{\textbf{Data-driven}\\\textbf{Baselines}}} 
& SFT with refusal trajectories & 0 & - & - & 0.697 & 0.000 & 0.697 & 0.697 \\
& & 10 & - & - & 0.692 & 0.041 & 0.722 & 0.706 \\
& & 20 & - & - & 0.679 & 0.059 & 0.722 & 0.700 \\
& & 30 & - & - & 0.664 & 0.040 & 0.692 & 0.678 \\
\cmidrule{2-9}
& RL from scratch (Unanswerable) & 0 & - & - & 0.620 & 0.002 & 0.621 & 0.621 \\
& & 10 & - & - & 0.644 & 0.040 & 0.671 & 0.657 \\
& & 20 & - & - & 0.580 & 0.185 & 0.712 & 0.639 \\
& & 30 & - & - & 0.453 & 0.367 & 0.716 & 0.555 \\
\cmidrule{2-9}
& Cold-start RL (Unanswerable) & 0 & - & - & 0.660 & 0.000 & 0.660 & 0.660 \\
& & 10 & - & - & 0.665 & 0.000 & 0.665 & 0.665 \\
& & 20 & - & - & 0.676 & 0.002 & 0.677 & 0.677 \\
& & 30 & - & - & 0.688 & 0.003 & 0.690 & 0.689 \\
\midrule

\multirow{9}{*}{\shortstack[l]{\textbf{Static Reward}\\\textbf{Baselines}}}
& RL from scratch ($r_{ref}$) & - & 0.00 & - & 0.636 & 0.012 & 0.644 & 0.640 \\
& & - & 0.05 & - & 0.000 & 0.999 & 0.000 & 0.000 \\
& & - & 0.10 & - & 0.000 & 0.999 & 0.000 & 0.000 \\
& & - & 0.20 & - & 0.000 & 0.942 & 0.000 & 0.000 \\
\cmidrule{2-9}
& Cold-start RL ($r_{ref}$) & - & 0.00 & - & 0.674 & 0.003 & 0.676 & 0.675 \\
& & - & 0.20 & - & 0.657 & 0.121 & 0.747 & 0.699 \\
& & - & 0.40 & - & 0.627 & 0.203 & 0.787 & 0.698 \\
& & - & 0.60 & - & 0.586 & 0.283 & 0.817 & 0.683 \\
& & - & 0.80 & - & 0.578 & 0.308 & 0.835 & 0.683 \\
\midrule

\multirow{12}{*}{\shortstack[l]{\textbf{\texttt{AWA-RL}}\\\textbf{(Ours)}}}
& RL from scratch ($\gamma$) & - & - & 1.000 & 0.443 & 0.306 & 0.638 & 0.523 \\
& & - & - & 0.500 & 0.514 & 0.311 & 0.746 & 0.609 \\
& & - & - & 0.250 & 0.578 & 0.214 & 0.735 & 0.647 \\
& & - & - & 0.200 & 0.579 & 0.187 & 0.712 & 0.639 \\
& & - & - & 0.125 & 0.624 & 0.082 & 0.680 & \textbf{0.651} \\
& & - & - & 0.000 & 0.636 & 0.012 & 0.644 & 0.640 \\
\cmidrule{2-9}
& Cold-start RL ($\gamma$) & - & - & 1.000 & 0.614 & 0.241 & 0.809 & 0.698 \\
& & - & - & 0.500 & 0.641 & 0.182 & 0.784 & 0.705 \\
& & - & - & 0.250 & 0.669 & 0.116 & 0.757 & 0.710 \\
& & - & - & 0.200 & 0.684 & 0.097 & 0.757 & \textbf{0.719} \\
& & - & - & 0.125 & 0.694 & 0.056 & 0.735 & 0.714 \\
& & - & - & 0.000 & 0.674 & 0.003 & 0.676 & 0.675 \\
\bottomrule
\end{tabular}%
}
\caption{Detailed experimental results on the \textbf{HotpotQA} dataset across all training paradigms and hyperparameters. We report Accuracy (Acc), Refusal Rate (Refuse), Precision (Prec), and Reliability-Aware F1 (\texttt{RA-F1}). The best performing configurations for \texttt{AWA-RL} in both from-scratch and cold-start settings are highlighted in bold. Dashes (-) indicate not applicable.}
\label{tab:hotpotqa_results}
\end{table*}

\end{document}